\documentclass{ieeeaccess}
\usepackage{cite}
\usepackage{amsmath,amssymb,amsfonts}
\usepackage{algorithmic}
\usepackage{graphicx}
\usepackage{textcomp}
\usepackage{array}
\usepackage{booktabs} 
\usepackage{longtable}
\usepackage{bm}
\usepackage{graphicx}   
\usepackage{amsmath}    
\usepackage{hyperref}   

\usepackage{float}
\usepackage{pdflscape}
\makeatletter
\AtBeginDocument{\DeclareMathVersion{bold}
\SetSymbolFont{operators}{bold}{T1}{times}{b}{n}
\SetSymbolFont{NewLetters}{bold}{T1}{times}{b}{it}
\SetMathAlphabet{\mathrm}{bold}{T1}{times}{b}{n}
\SetMathAlphabet{\mathit}{bold}{T1}{times}{b}{it}
\SetMathAlphabet{\mathbf}{bold}{T1}{times}{b}{n}
\SetMathAlphabet{\mathtt}{bold}{OT1}{pcr}{b}{n}
\SetSymbolFont{symbols}{bold}{OMS}{cmsy}{b}{n}
\renewcommand\boldmath{\@nomath\boldmath\mathversion{bold}}}
\makeatother

\def\BibTeX{{\rm B\kern-.05em{\sc i\kern-.025em b}\kern-.08em
    T\kern-.1667em\lower.7ex\hbox{E}\kern-.125emX}}

\begin{document}
\history{Date of publication xxxx 00, 0000, date of current version xxxx 00, 0000.}
\doi{10.1109/ACCESS.2024.0429000}

\title{A Study on the Application of Explainable AI on Ensemble Models for Predictive Analysis of Chronic Kidney Disease}

\author{\uppercase{K M Tawsik Jawad}\authorrefmark{1}, 
\uppercase{Anusha Verma}\authorrefmark{2} \IEEEmembership{Student Member, IEEE}, Fathi Amsaad\authorrefmark{3},
\IEEEmembership{Senior Member, IEEE} and Lamia Ashraf \authorrefmark{4}}

\address[1]{Department of Computer Science, University of Cincinnati (e-mail: jawadkk@mail.uc.edu), Cincinnati, OH USA }
\address[2]{Department of Computer Science, Wright State University, Dayton, OH USA (e-mail: verma.39@wright.edu)}
\address[3]{Assistant Professor, Department of Computer Science, Wright State University, Dayton, OH USA (e-mail: fathi.amsaad@wright.edu)}
\address[4]{Intern Doctor, Tangail Medical College and Hospital, Tangail, Bangladesh (e-mail: lamiya.ashraf456@gmail.com) }

\markboth
{Author \headeretal: Preparation of Papers for IEEE TRANSACTIONS and JOURNALS}
{Author \headeretal: Preparation of Papers for IEEE TRANSACTIONS and JOURNALS}

\corresp{Corresponding author: K M Tawsik Jawad (e-mail: jawadkk@mail.uc.edu)}

\begin{abstract}
Chronic Kidney Disease (CKD) is one of the widespread Chronic diseases with no known ultimo cure and high morbidity.  Research demonstrates that progressive CKD is a heterogeneous disorder that significantly impacts kidney structure and functions, eventually leading to kidney failure. The goal of this research is to first develop an accurate ensemble model for prediction of unseen cases of CKD given the biomarkers. Also, we have implemented the Explainable AI (XAI) algorithms to interpret the decision-making process of the ensemble models in terms of dominating features and the feature values. The takeaway from our research is to aid the physicians make an informed decision about the disease and provide a case by case explanation behind their decisions. Also, XAI algorithms would allow the patients or subjects understand the causes behind their disease at early stages so that they can be cautious about the progression of the disease to later stages.
\end{abstract}


\titlepgskip=-21pt

\maketitle

\section{Introduction}
\label{sec:introduction}
The clinical definition of CKD is based on the presence of kidney damage or Decreased Kidney Function. That is when Glomerular Filtration Rate (GFR) less than 60 mL/min per 1.73 m$^2$ for a period of 3 months or more. 

\subsection{Analysis of biomarkers for CKD}
The risk factors of occurrence for CKD increase manifolds in the patients suffering from Diabetes Mellitus or Hypertension. Causation of the disease and management are measured in terms of stages of disease severity and clinical diagnosis. Routine Laboratory tests specifically Blood and Urine tests reveal the occurrence of CKD. The progression of the disease is also measured by GFR quantity. Management of this disease needs change in dietary plans and lifestyle modifications. But for all this to happen, it is highly important that the disease is detected beforehand. There are some factors that leads to kidney diseases like presence of Type I or Type II Diabetes, Presence of polycistic or other inherited kidney diseases, prolonged obstruction of urinary tract, recurrent kidney infection etc. The first sign of kidney disease from Diabetes is protein in the urine. When the filters are damaged, usually a protein named Albumin passes into the urine from the blood. But a healthy kidney will not let this occur. High blood pressure can cause CKD and can happen as a result of CKD as well \cite{b1}. CKD is a major health concern that affects millions of people around the world every year. The irreversible nature of this disease not just leads to comorbid diseases like Diabetes Mellitus, Hypertension etc. but also it can permanently damage the kidney by progressing to Acute Kidney Injury (AKI) or End Stage Renal Diseases (ESRD). Unlike the healthcare facilities available at developed countries, in the developing countries, the CKD is not identified till the ESRD occurs where dialysis remains the only option. The common practice associated with identification of CKD is if the Glomerular Filtration Rate (GFR) quantity is less than 60 mL/min or presence of biomarkers that indicate kidney damage for last 3 months. In most of the clinical practices, common tests for CKD includes measurement of GFR. GFR quantity is calculated from the serum creatinine concentration. From the Urine test, albuminuria or albumin to creatinine ratio is indicative of CKD \cite{b3}.

\subsection{CKD as a Health Concern}
CKD is a severe public health problem that affects general population of all ages. According to reports of Centers of Disease Control and Prevention (CDC), about 37 Million adults in USA are unaware about having CKD. In the US, approximately 360 people start dialysis treatment for kidney failure every hour. Almost three out of four new cases of kidney failure occurs due to high blood pressure and diabetes which makes them leading causes of kidney failure in the United States \cite{b2}. According to reports of Mayo Clinic, CKD could also lead to Anemia, Fluid Retention, High Blood Pressure, Weak Bones, Risk of Bone Fractures, Pregnancy Complications and cause decreased Immunoresponse \cite{b3}. With CKD, the kidneys of patients cannot make enough Erythropoietin which causes drop in red blood cells and gradually leads to Anemia.  Kidneys are the primary organ for filtering waste materials. When that organ loses it's capacity of filtering, naturally fluid overload, electrolytes and wastes are built up on the body. Furthermore, CKD causes the blood vessels to get narrowed which increases Blood Pressure. CKD causes vitamin D synthesis leading to Weak Bones\cite{b3}. More symptoms of CKD are Nausea, Vomiting, Loss of Appetite, Muscle cramps, Swelling of Feet and Ankles, Shortness of Breath etc. It is hard to identify between the causes and complications of CKD without clinical diagnosis and follow ups. Because complication of CKD can also be a cause and vice versa.

\subsection{CKD in Monetary Numbers}
Tackling chronic kidney diseases has many economical challenges also. According to reports from American Society of Nephrology, approximately 26 million Americans have some evidence of CKD and are at risk to develop kidney failure. Another 20 million are at risk of developing the kidney diseases. The number of people diagnosed with CKD has doubled in the past two decades and older Americans in the age range of 60-70 are more at risk of developing this disease. The progression of kidney disease eventually leads to End Stage Renal Diseases (ESRD) and the annual cost of treating ESRD is currently over \$32 billion. To put things more into global perspective, CKD ranks 12th as a cause of death and 17th as a cause of disability \cite{b4}.
With the state of the art healthcare facilities in the West, less reports are obtained for developing countries in the South-East Asia. In the public hospitals of the Indian sub-continent, treatment of ESRD has become a low priority. With the absence of health insurance plans, less than 10\% of ESRD patients receive any Renal Replacement Therapy. Treatment gets stopped for vast majority of the patients in first 3 months due to the overwhelming cost of the process\cite{b4}.

\subsection{Difficulty in CKD Analysis from Engineering Perspective}
Getting our hands on healthcare data for a particular domain in itself is a difficult task. Even if those datasets are obtained, it is important to be careful about cognitive and unconscious biases in these data \cite{b5}. After diligently searching for patient data in CKD occurrence and progression, it was identified that very few publicly available datasets are there for CKD patients. To gain access to real time patient data, it was needed to resort to hospitals and clinics in the locality which needed proper channels and authorization. Because hospital authorities have to maintain confidentiality of their patients. Ideally, all demographic, clinical and socio-economic factors required for distinguishing the CKD with non-CKD patients would be a strong dataset to work on this domain. Dataset selection was concluded after going through number of research papers, nephrology journals and collaborating with the university biology department. It was found the dataset on CKD from UCI Machine Learning repository was close enough for our research requirements. Although this dataset does not have any diversity in terms of ethnicity, gender or any socio-economic conditions and also it lacked in the clinical features required for distinguishing CKD patients, this is the closest dataset obtained for this research keeping the required features in mind after evaluating all other datasets.

\subsection{A Brief on Existing Research on CKD}
Existing works on prediction and diagnosis of complications associated with CKD use several pre-processing techniques for effective feature elimination. Ahmed J, Dhiya et al have used Random Forest, Support Vector Machine and Logistic Regression to find out optimal set of parameters for prediction of CKD.  To find out the capacity of contributing features in identifying the risk factors of CKD, the scope of this research lies in application of Explainable AI (XAI) in explaining the decision-making process of ensemble models. The primary research question is what factors with what values contribute towards making CKD or Non-CKD prediction. Moreover which models would be better suited in terms of accuracy metrics and interpretability measures in this task. The goal of this research would be to assist the clinicians as end users in explaining risk factors in CKD progression. The clinicians in turn, can inform the patients about diet, lifestyle changes to prevent the progression to next stages. Dataset was collected on 400 subjects where some of them had CKD and others were healthy individuals from Apollo Hospital, India. The current scope of this research has also been compared with existing research in terms of Accuracy, Interpretability and Fidelity measures. The research findings were validated and also corrected by nephrologists for further research directions.  Jiongming Qin et al at has used KNN imputation method to generate 5 more datasets out of the UCI ML dataset applying the ML models \cite{b4}. Along with the biological factors, they have found strong correlation of a demographic factor like age in making a distinction between CKD and Non-CKD patients. Also, to find out optimal subset of features they employed Gini Index in Random Forest which determined the feature importance values at each trees. For decision trees and random forests, the algorithms are designed in a way that the features possessing the least value of the Gini Index would get preferred by these algorithms. This particular technique is employed to calculate the predictability of a particular feature to the response variable in\cite{b3}.

\subsection{Use of XAI in CKD}
Among existing literature on the applications of XAI on CKD, studies conducted by Manju et al at \cite{b66} have used traditional pre-processing techniques like min max scaling and Z-score standardization to keep all the features in a range. Z-score standardization usually would transform the values of features so that the standard deviation of features would be 1 with the mean value 0. If x is the actual value of the feature and x' is the transformed value of the feature after Z-score standardization, then

\begin{equation*} x^{\prime}=\frac{x-mean(x)}{std(x)} \tag{2}\end{equation*}

Equation 2 would provide us with the transform equation for Z-score standardization. In the studies conducted by Manju et al, they have used Gini Index as the impurity measure for branching out leaf nodes from the root nodes. The higher the value of the feature by the Gini Index, the more importance the Decision Tree model is providing to the features. The Gini Index measure is also calculated based on the proportion of the samples that reach the leaf nodes.

\subsection{Overview of Our Research}
\begin{figure*}[!t]
    \centering
    \includegraphics[scale=0.59]{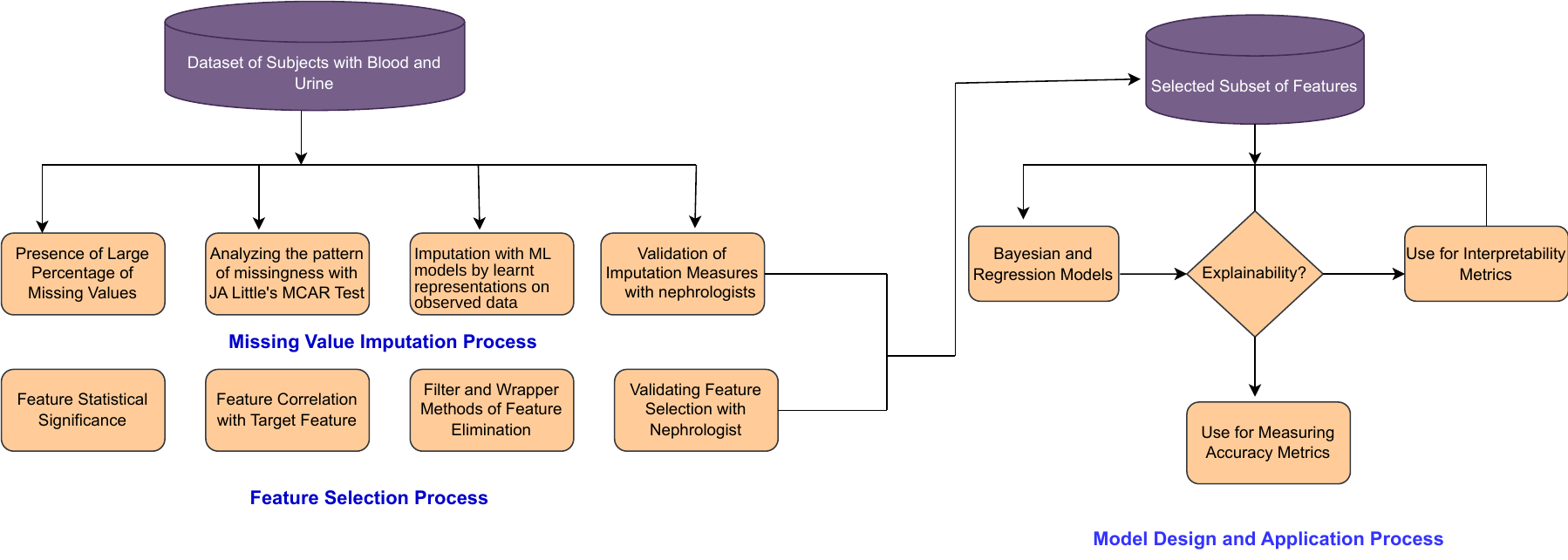}
    \caption{Workflow Diagram of Overall Research Process}
    \label{Fig:model}
\end{figure*}

Figure 1 represents the overall workflow of the research where the dataset collected from the blood and urine tests go through the feature selection process following missing value imputation. Next, the selected subset of features is fed to Bayesian and Ensemble models. The model design and application process is followed with the XAI algorithms and finally comparing the scope of the research with existing interpretable models on CKD.

The following section after the introduction would describe the literature review of causes and implications of CKD, existing scope of work with CKD and analysis of missing value imputations. Following that section, the dataset specification section describes the feature distribution and the dataset pre-processing. After these, the next sections are designed towards model selection and implementation and experiment design, parameter tuning of the models. Finally, the results from the experiment sections are analyzed, interpreted and compared with the existing research on the applications of XAI on CKD.

\section{Background Study}

The Background Study of this research is designed maintaining a flow to the research objective in question: Analyzing the critical features that contribute towards prediction of CKD in subjects. The goal is identification of these features beforehand so that the patients are well aware of their critical body vitals helping them take better decisions in prevention of CKD towards End-Stage Renal Disease (ESRD).

\subsection{Dealing with Healthcare Datasets}
 Most of the clinical studies are longitudinal cohort studies where the subjects are regularly followed up after certain intervals. It is natural in these datasets to have missing values in certain features as many subjects from these studies chose not to follow up anymore. Moreover, some cross-sectional clinical data are prone to human error on the data entry. The missing data occurrence in these studies add more complexity to the already complicated domain of healthcare. Several studies conducted by recently have used iterative imputation techniques to account for missing value imputation, and used naive supervised learning techniques to predict CKD cases validated by cross validations on the test dataset \cite{b5}. Among other studies specially on the CKD diagnosis from patient data from Brazil, they addressed some critical aspects specially for early detection of CKD \cite{b6}. In the developing countries like Brazil, it becomes very important to aid the physicians from an automated system because in the clinical settings, Kidney Diseases are not detected specially till they reach towards ESRD. The requests for the clinical datasets have to go through proper channels, authorities. The researchers requiring these datasets also have to follow the code of compliance dealing with human subjects and perform training procedures before diving into research with the clinical data.

In the exploratory data analysis stage of \cite{b6}, augmentation of their dataset was validated by a nephrologist having 30 years of experience in the field of CKD treatment, detection and prognosis. The data augmentation process was manually validated by an experienced nephrologist and also they performed automated augmentation using Synthetic Minority Oversampling Technique
(SMOTE). The oversampling techniques were applied because their study contained a very small and imbalanced dataset.

Studies conducted by Vintrella et al have used dataset from hospitals to measure the progression of CKD stages \cite{b7}. Their fundamental research question was finding out the time remaining for dialysis of a patient suffering from CKD. They needed to distinguish between patients suffering from CKD versus patients having Acute Kidney Injury (AKI) in the g4 stage of the CKD. Collecting the dataset required them to collect unstructured information from texts like family history of diabetes, anemia, hypertension, presence or absence of kidney stones, smoker or not, any renal transplant or not etc. Their studies also included both classification and regression approach where they treated the time to start dialysis as the target feature.

\subsection{Application of XAI in CKD Research}
While significant studies have focused on application of Machine Learning models on several different use cases of CKD prognosis, diagnosis and treatment, the area of Explainable AI is still quite under-explored. It becomes imperative to intepret the decisions of the ML models since those decision would need to be validated by nephrologists as end users. Most of the research methodologies primarily focus on feature based XAI techniques. Studies conducted by Pedro A. Moreno-Sánchez \cite{b8} have used interpretability, fidelity measures to make a comparative study of their proposed approach with other concurrent applications. Most of the models applied by them are ensemble tree based models. They have applied XAI feature based techniques to extract more contributing features which would be fed to these tree based models for performance improvement.

In more critical domains such as kidney image analysis for discovering cysts, the explainable AI was used to identify specific regions of a kidney image that deep learning models use to distinguish between classes \cite{b9}. Extensive studies conducted by Gabriel et al have collected demographic information from a sample 20,000 patients with CKD and same sample sized healthy individuals for a Case Based Analysis (CBR) approach \cite{b10}. Their feature sets of 7,493 variables combined of all demographic information and diagnosis of previous CKD occurrence. CBR approach basically retrieves the most similar cases to the case in query and an adaptation stage that would combine the similar cases to build a solution to the case in query. The Neural Network Case Based Reasoning (NN-CBR) method is used so that explanations provided to the user is based on the comparison of the input of the Neural Nets to similar examples which are obtained by the CBR system. Out of the huge variable set, 15 variables were identified as dominant features by the feature importance scores of Random Forest classifier. Among other analysis, they identified the age range where the CKD occurence peak and plateau out, and also identified the risk of developing CKD according to places of residence by dividing the number of persons identified with CKD in each department with the number of persons in that department. Among the feature sets age, essential hypertension, low back pain, urinary tract infection, pain in joint were the top 5 features for explanatory model.

Among other XAI applications in different domains of healthcare, studies conducted by Devam Dave et al use example based techniques along with feature based techniques \cite{b11}. They used Anchors which generate local regions that denotes a better construction of the dataset for explanation and also have less computational expense than SHAP. Anchors were used to identify as ``if then" conditionals with independent feature values as if condition and the dependent feature value as ``then" statement. The Counterfactuals were employed to evaluate what sort of changes would need to occur in the independent features so that the prediction of the model is switched to the contrasting class. Counterfactuals with prototypes provide similar but faster explanations on the basis of prototype or samples that is representative of the instances belonging to a class. In the contrastive explanation models, Pertinent Positives work similar to Anchors where Pertinent Negatives work similar to Counterfactual \cite{b12}. The concept of integrated gradients from \cite{b13} has been explained pretty well from the MNIST Fashion dataset at \cite{b11}. For the Integrated gradients, positive attributions are the ones that make a positive influence for the model to identify it as the positive class and negative attributions are the opposite ones. They provided the example with ``Shoes" being the positive class where pixel values of features like laces and shoes were the positive attributions and some features of shoe laces contributed towards identifying it as ``Not a Shoe" or the negative class. Tables 1 contains different techniques employed by authors of the background study with their pros and cons. Table 2 contains a more detailed analysis of the specific dataset used, experiment strategies, models used and the accuracy metric results.

\clearpage
\begin{table}[!h]
\centering
\caption{Comparison of Techniques for CKD Prediction}
\begin{tabular}{|m{2cm}|m{3cm}|m{4cm}|m{4cm}|} 
\hline
\textbf{Authors} & \textbf{Techniques} & \textbf{Pros} & \textbf{Cons} \\ \hline
Silveira et al \cite{b6} & 
Data augmentation by consultation with nephrologist, SMOTE. &
Both techniques are used for handling class imbalanced dataset. & 
Use of Grid Search Cross Validation to find best set of parameters for the ensemble models increasing time complexity. \\ \hline

Vintrella et al \cite{b7} &
Collection of patient’s demographic information. &
Family history information gathering like family history
of diabetes, anemia, hypertension, presence or absence of
kidney stones, smoking etc. &
Research analysis limited to only Caucasian ethnicity. They have not used features like Body Mass Index (BMI), Protein Urea, Diet etc. which are important contributors in CKD prediction. \\ \hline

Pedro et al \cite{b8} &
XAI feature based techniques on ensemble models. &
Calculation of interpretive measures like Interpretability (I), Fidelity (F).  &
Lack of validation data as research is limited to analysis of one CKD dataset from UCI ML repository. \\ \hline

Bayram et al \cite{b9} &
Use of XAI on Kidney Image Dataset. &
Distinction between different kidney diseases like kidney cyst, kidney stone by XAI. &
Incorrect predictions were not analyzed, dataset was limited to only one ethnicity. \\ \hline

Gabriel et al \cite{b10} &
Case Based Reasoning (CBR) on 20,000 patients with 7,493 variables encompassing demographic information and previous diagnosis of CKD. &
Neural Net Based CBR was employed to process the explanation of the CKD decision based on similar examples seen earlier. &
Selected features in their experiments were binary, they have not worked on any continuous variable. User feedback from the explanatory models is not implemented.\\ \hline

Devam et al \cite{b19} &
Example Based XAI techniques on Heart Disease Dataset. &
Use of Anchors as “If-then” statement and Counter-factuals to understand the changes need to happen for each feature to predict the contrasting class. &
Comparative Analysis not found with concurrent datasets on Heart Disease or other medical datasets. \\ \hline

Our methods &
Missing data analysis by JA Little's MCAR Test, Imputation by ML models and nephrologists, Feature and Example based XAI applications. &
Thorough analysis of missing data, precise feature selection methods, patient specific contrastive class prediction details, rigorous pre-processing of dataset, application of novel fidelity measures on this CKD dataset, approaches of building a more clinically comprehensive  CKD dataset.&
No discussions on demographic variables, no calculation of internal fidelity, less analysis of inter-feature correlation in independent features. \\ \hline

\end{tabular}

\label{tab:ckd_comparison}
\end{table}

\clearpage
\vspace{-0.5cm}
\begin{landscape}
\begin{table}[!htbp]
\small
\centering
\caption{Experimental strategies and models in CKD prediction studies.}
\label{tab:ckd_studies}
\begin{tabular}{|p{2cm}|p{4cm}|p{5cm}|p{3.5cm}|p{1.2cm}|p{1.2cm}|p{1.2cm}|p{1.2cm}|}
\hline
\textbf{Study Title} & \textbf{Dataset Used} & \textbf{Experiment Strategy} & \textbf{Models Used} & \textbf{Accuracy} & \textbf{Precision} & \textbf{Recall} & \textbf{F1 Score} \\ \hline
Silveira et al \cite{b6} & Medical records of Brazilians with or without CKD, having attributes like hypertension, diabetes, creatinine, etc. & Hold-out validation, multiple stratified cross-validation (CV), and nested cross-validation. & Decision Tree (DT), Random Forest, and Multi-Class AdaBoosted DTs & 0.95 & 0.92 & 0.93 & 0.94 \\ \hline
Vintrella et al \cite{b7} & Data extracted from the Vimercate Hospital EMR with features like GFR over a time period, diabetes, chlorine, cardiopathic, anemic, etc. & Classification and regression approaches to determine exactly when patients would need dialysis. For classification metrics, 10-fold cross-validation was used to analyze accuracy metrics. For regression, Mean Squared Error (MSE) was used to estimate prediction error. & Logistic Regression, Decision Tree (DT), Random Forest, Fully Connected Neural Nets & 0.95 & 0.92 & 0.93 & 0.94 \\ \hline
Pedro et al \cite{b8} & CKD dataset from the UCI ML Repository. & 5-fold cross-validation used to measure unseen test set predictions, with different interpretability measures calculated using permutation feature importance techniques. & Random Forest, Decision Tree (DT), AdaBoost, Extra Trees Classifier & 0.95 & 0.92 & 0.93 & 0.94 \\ \hline
Bayram et al \cite{b9} & The kidney dataset containing images of patients diagnosed with kidney cysts, normal, and stone findings. & A supervised image classification task using deep learning models. & You Only Look Once (YOLO v7) and YOLO v7 Tiny architectures & 0.85 & 0.88 & 0.83 & 0.85 \\ \hline
Gabriel et al \cite{b10} & RIPS database collected from the Colombian Ministry of Health and Social Protection that contains individual record of health service delivery. & Neural network with hyperparameters tuned on the dataset. To tackle overfitting, early stopping and dropout were applied as regularization techniques. & Neural Network, Support Vector Machine (SVM), and Random Forest (RF) & 0.95 & 0.94 & 0.97 & 0.95 \\ \hline
Devam et al \cite{b19} & The heart disease dataset from Cleveland UC Irvine dataset. & Feature-based and example-based XAI algorithms are applied on the ensemble models for prediction analysis. & Logistic Regression (LR), XGBoost; SHAP, Counterfactuals, Anchors, Integrated Gradients applied on LR and XGBoost & N/A & N/A & N/A & N/A \\ \hline
\end{tabular}
\vspace{-0.5cm}
\end{table}
\end{landscape}

\subsection{Addressing Data Security in AI Applications}
In terms of data security, several challenges occur like network attacks, system bias, mismatch attacks etc. that are specially attributed to the black box nature of the AI models \cite{b14}. The potential solution to these attacks have been approached with XAI techniques in \cite{b15, b16}. Not only security aspect is concerning, there are ethical aspects of applying AI models in clinical domains. It is highly critical to develop responsible practices of AI applications as patient health, hospital staff trust everything becomes at a stake based on the applications \cite{b17}. Moreover, the Machine Learning models are not free of different kinds of bias. Bias could happen in the process of collection of data from human end like selection bias, measurement bias, representation bias. On the application ends, there could be algorithm bias of choosing particular models to suit particular types of datasets. As ML practitioners, there could be confirmation bias where engineers and researchers applying the models have some pre-conceived notion about the results and would want to keep the positive findings on those biases \cite{b18}. Therefore, it becomes highly imperative to use responsible AI models that can tackle the various kinds of bias in the dataset and algorithm levels. This is where careful investigation of the AI models is required before deploying to personalized health challenges.

\begin{table}[!t]
\centering
\caption{Percentage of Missingness in all columns}
\begin{tabular}{|c|c|}
	\hline
	\textbf{Feature} & \textbf{Percentage of Missingness}\\
	\hline
	Age & 2.23 \\
	\hline
    Bp & 2.98 \\
	\hline
	Sg & 11.70 \\
	\hline
	Al & 11.44 \\
	\hline
    Sg & 12.18 \\
    \hline
    Rbc & 37.81 \\
    \hline
    Pc & 16.16 \\
    \hline
    Pcc & 0.99 \\
    \hline
    Ba &  0.99 \\
    \hline
    Bgr &  10.94 \\
    \hline
    Bu &  4.72 \\
    \hline
    Sc &  4.22 \\
    \hline
    Sod &  21.64 \\
    \hline
    Pot &  21.89 \\
    \hline
    Hemo &  12.93 \\
    \hline
    Pcv &  17.66 \\
    \hline
    Wbcc &  26.36 \\
    \hline
    Rbcc &  32.58 \\
    \hline
   Htn &  0.50 \\
    \hline
    Dm &  0.50 \\
    \hline
    Cad &  0.50 \\
    \hline
    Appet &  0.24 \\
    \hline
    Pe &  0.24 \\
    \hline
    Ane &  0.24 \\
    \hline
\end{tabular}

\end{table}

\section{Dataset Pre-Processing}
The dataset is collected from UCI Machine Learning Repository on Chronic Kidney Diseases from Apollo Hospital, India. The dataset has 25 features out of which 14 are nominal and 11 are numeric. The dataset is generated using the blood and urine test results of 400 subjects. 250 of these subjects were diagnosed with CKD and the rest 150 were healthy individuals. The dataset pre-processing is performed by analyzing the missing values, imputation of missing values, analyzing statistical measures on feature significance, finding out the correlation of independent features with the target feature and employing various feature extraction and elimination techniques.

\subsection{Missing Values}\label{AA}
Since the dataset is collected from manual input of the patients vitals and results of blood tests, naturally there are lots of missing values in the dataset. From table I, it is observed that 12 out of the 25 total features had moderate to high amount of missing values.

Some features like `Rbc', `Sod', `Pot', `Wbcc', `Rbcc' has high percentage of missing values (`High' here indicates more than 20\% of missingness). Features like `Htn', `Dm', `Cad', `Appet' have the least amount of missing values. The missing values were not ``NaN" as found on maximum datasets, rather it was ``?" in significant number of rows. After accessing the dataset, missing values were found to be completely at random (MCAR distribution).  No pattern was found for the missing values neither any relation was found for the missing values to the actual values in the dataset. This randomness of missing values would give us an advantage to perform statistical analysis on them and the analysis results would be unbiased.

Several consolidated research domains on addressing the patterns and problems on missing data in clinical studies have concluded three patterns of missing data. First of them is the Missing Completely at Random  (MCAR) distribution. In this distribution, the missing data does not depend on the observed data or the missing data themselves. The second pattern is Missing at Random (MAR) distribution. In this pattern, the missing data is related to the observed data but not related to the unobserved or missing data in the dataset. The final distribution is the Missing Not at Random (MNAR) distribution.

\begin{table}[!t]
\centering
\caption{RJA Little's MCAR Test On Different Sample Sizes}
\begin{tabular}{|p{1 cm}|p{1 cm}|p{2 cm}|p{1cm}|}
	\hline
	\textbf{Sample Size} & \textbf{Statistic} & \textbf{P-Value} & \textbf{Missing Patterns}\\
	\hline
    120 &   1121 & 0.0000223 & 48\\
	\hline
	240 &   1794 &  0.00000000310 & 73\\
 \hline
 320 & 2010 & 2.44e-15 & 78\\
\hline
\end{tabular}

\end{table}  

If missingness is related to data or information that is not available, then the distribution of missingness is not at random, it becomes difficult to impute the missing values in a ``MNAR" distribution \cite{b62}. If the entire dataset is considered to be matrix Y, it can be decomposed into  Y$_{o}$ and Y$_{m}$, which denote the observed and missing data. Let R denote a missing value matrix defined by the following probability: R = 0 if Y is observed and R = 1 if Y is missing. Let us consider q to be a vector of values that indicate the association between missingness in R and the dataset Y. Hence, the The probability of MCAR would be defined as: \(p(R|q)\).  The research process of finding out the pattern of missingness is first looking at the dataset and then following the process of mathematical induction of concluding the pattern of missingness. Looking at the dataset in the data pre-processing stage, this study assumed the ``Null Hypothesis" that the missing data is MCAR. 

Experiments were designed to analyze whether this null hypothesis could be rejected or not. This research used the extensive test conducted by RJA Little at \cite{b63} where the significance level, degrees of freedom and test statistic P-value would be analyzed to find whether the null hypothesis could be rejected or not. The significance level was determined by Fisher's threshold \cite{b64}. The Little's MCAR Test needs all the variables to be numerical so the categorical variables were converted to numeric ones and then randomly 30\%, 60\% and 90\% samples were chosen for the test. 

Looking at the MCAR Test results on different sample sizes from table II, the P-Values are all less than Fisher's threshold of 0.005. As more samples are selected, the P-Values are decreasing and moving towards 0. But it is also noticable that, the statistic value is increasing with the number of missing patterns. Since in all of the sample sizes chosen here have less than Fishers' threshold of 0.005, so Null Hypothesis can't be rejected. The increasing value of the statistic with the small p-values infer that the missing value distribution is not completely at random (MCAR). There is not sufficient evidence against the null hypothesis that missing value distribution is MCAR. That also implies that missing value distribution could be MNAR or MAR. So, deleting the missing values here is not an effective measure since deletion of those rows would result in loss of critical information about the patients. That's why this research focused more on the imputation techniques of these values.

\begin{table}[!t]
\centering
\caption{Highly Correlating Features with Target Feature}
\begin{tabular}{|c|c|}
	\hline
	\textbf{Feature Name} & \textbf{Co-relation Value} \\
	\hline
	Hemo & 0.76\\
	\hline
    Sg & 0.73\\
	\hline
	Pcv & 0.71\\
	\hline
    Rbc & 0.67\\
    \hline
	Al & -0.65\\
	\hline
    Htn & -0.59\\
    \hline
    Dm & -0.56\\
\hline
\end{tabular}

\end{table}

\subsection{Imputation of Missing Values}\label{AA}
The dataset had a significant percentage of missing values in a number of columns. Previous approaches of mean, median or KNN imputations were discarded because it could lead to data leakage issue of the subjects. Moreover, in a sensitive issue like medical datasets, just using traditional means of imputation is not quite clinically feasible. So, a more effective measure was using training-testing split of the original dataset to impute the missing values. First we found out the exact rows where none of the features had any missing values and used that particular dataset as the training dataset. Training was performed with a linear regression model as we had to predict the continuous values in some features. Then we did some pre-processing like cleaning strings, tabs, whitespaces on the test dataset with the missing values. One key issue in the missing data imputation was it was difficult to select a target feature and treat other features as independent in the test dataset. This happened because whichever feature we chose as dependent, there were more than a couple of independent features which had missing values. This is where we had to sort the features in the test dataset by choosing one feature as target feature for imputation. If that target feature was termed as "A", we looked at which are the exact features that had values present corresponding to the missing values of feature "A". Then in the training dataset, we just trained on those set of features to impute the values in feature "A". Similar process was followed for imputation of all the features that had missing values. Some post-processing was also required to organize the imputed values in the final imputed dataset. The data imputation process was also validated by the nephrologists Dr. Md. Nabiul Hassan Rana from Asgar
Ali Medical College, Dhaka, Bangladesh and Dr. Nazneen Mahmood from Anwar Khan Modern Hospital, Dhaka. Both of them agreed that the mathematical mean, median imputation on clinical data could lead to wrong calculations and introduce bias in the models. So, after a collective discussion, they also agreed that the best way to impute the missing data in this clinical setting would be to look at the observed values, learn the representations and then use the learnt representations to get an estimate of the missing data.

\subsection{Statistical Measures on Feature Significance}\label{AA}
 To find out which features out of the 25 features on the blood test and urine tests of the patients are statistically significant, a Logit model was fitted on the dataset. After analyzing the coefficients and the P-Values found from the fitted Logit Model, 7 features have been deemed to be statistically significant ones by the model.

From table IV, Only those features are deemed as statistically significant in CKD prediction where the `P values' are less the Fisher threshold of 0.005. The same process of ``Null Hypothesis" testing is also applied here only in this case the Null Hypothesis refers to the hypothesis that features are strongly significant in the prediction of the target ``CKD" feature. Or in other words, the relationship between each of the feature with the target feature is not due to random chance. Looking at the `P values' in table IV, it is evident that all the 7 features denoted here are indicative of the acceptance of the ``Null Hypothesis" and rejecting any alternative hypothesis.

\begin{table}[!t]
\centering
\caption{Statistically Significant Features as Found by the Logit Model}
\begin{tabular}{|c|c|c|}
	\hline
	\textbf{Feature Name} & \textbf{t-Value} & \textbf{P values }\\
	\hline
	Sg & 7.209 & 3.14e-12\\
	\hline
    Al & -6.527 & 2.18e-10\\
	\hline
	Rbc & 13.818 & 1e-16\\
	\hline
    Hemo & 3.468 & 0.00005\\
    \hline
	Bgr & -2.193 & 0.02891\\
	\hline
    Wbcc & -2.630 & 0.00888\\
    \hline
    Htn & -2.191 & 0.02909\\
\hline
\end{tabular}

\end{table}

\begin{table}[!t]
\centering
\caption{Feature Entropy and Information Gain Values}
\label{tab:feature_info_gain}
\begin{tabular}{|l|c|c|}
    \hline
    \textbf{Feature Name} & \textbf{Information Gain} & \textbf{Entropy} \\ \hline
    Hemo                 & 0.45                     & 7.01             \\ \hline
    Sg                   & 0.43                     & 3.01             \\ \hline
    Pcv                  & 0.42                     & 5.84             \\ \hline
    Rbcc                 & 0.42                     & 6.72             \\ \hline
    Sc                   & 0.42                     & 5.56             \\ \hline
    Al                   & 0.37                     & 2.80             \\ \hline
    Sod                  & 0.28                     & 5.81             \\ \hline
    Rbc                  & 0.26                     & 0.99
    \\ \hline
    Htn                  & 0.23                     & 0.99
    \\ \hline
    Pot                 & 0.21                      & 5.80
    \\ \hline
\end{tabular}
\end{table}

\begin{table}[!t]
\centering
\caption{Selected Set of Features from Filter-based and Wrapper Methods}
\begin{tabular}{|p{4 cm}|p{4 cm}|}
	\hline
	\textbf{Technique} & \textbf{Selected Features} \\
	\hline
	Information Gain &  `hemo', `pcv', `sg', `rbcc', `sc', `al', `sod', `rbc', `htn', `pot' \\
	\hline
    Variance Threshold & `age', `bp', `al', `su', `bgr', `bu', `sc', `sod', `pot', `hemo', `pcv', `wbcc', `rbcc'\\
	\hline
	Forward Selection & `age', `bp', `sg', `al', `rbc', `pcc', `ba', `pot', `hemo', `htn'\\
	\hline
    Recursive Feature Elimination & `al', `su', `rbc', `sc', `pot', `hemo', `rbcc', `htn', `dm', `pe'\\
\hline
\end{tabular}

\end{table}

\subsection{Correlation With the Target Feature}\label{SCM}
In order to validate whether the fitted logit models have correctly identified the statistically significant features, it was found out which features are more strongly correlating with the target class. For this analysis, a threshold of 0.5 as the cut off was chosen. So, in the Pearson Correlation matrix, the correlation values above that threshold are strongly correlating features.

So, from the correlation values, it is observed that most of the statistically significant variables deemed by the logit model are also highly correlated features as determined by the Pearson Correlation Matrix.

\subsection{Feature Selection Process}\label{SCM}
Since 25 features were initially present for the dataset, it is important to identify features would provide little information in terms of variability in the prediction. At first, several filter based methods of feature selection were applied, then the wrapper methods for the feature selection were employed. In this way, it was easier for us to determine which features are contributing more in the prediction of the CKD.

\begin{table}[!t]
\centering
\caption{Accuracy Metrics of Applied Machine Learning Models with Selected Features}
\begin{tabular}{|p{1 cm}|p{1 cm}|p{1 cm}|p{1 cm}|p{1 cm}|}
	\hline
	\textbf{Model} & \textbf{Precision } & \textbf{Recall } & \textbf{F-1 Score }& \textbf{Accuracy }\\
 \hline
	LR &   0.97 & 0.97 & 0.97 & 0.97\\
	\hline
    NV &   0.95 & 0.95 & 0.95 & 0.94\\
	\hline
	SVM &   0.98 & 0.98 & 0.98 & 0.98\\
	\hline
 Adaboost &   0.97 & 0.97 & 0.97 & 0.97\\
	\hline
  XGBoost &   0.98 & 0.98 & 0.97 & 0.98\\
  \hline
  DT &   0.98 & 0.98 & 0.98 & 0.96\\
\hline
  RF &   0.98 & 0.99 & 0.98 & 0.98\\
\hline
\end{tabular}

\end{table}

The information gain method was first employed which calculates reduction of entropy or which features retain more information in prediction of the target feature. The higher the value of entropy, higher is the variance within that feature. So, with information gain we are selecting only those features which have large to moderate entropy and also contribute towards prediction or reducing the determining the uncertainty of target feature as shown in table 5. Then the Variance Threshold method was applied to remove the constant and quassi-constant features. Constant features are the ones which have variance across the entire dataset. We can also say, they are homogenous features. Quassi-constant features are the ones where one value dominates 99\% of the records. We need to remove these constant and quassi-constant features because no variance in the feature set would only increase computational complexity of the models, not adding any information in the features.  A high threshold of 0.75 was selected as the criteria to remove any feature that has less than 0.75 as the variance. Both the forward and backward feature elimination yielded same set of variables, so only the forward selection results were presented. Logistic Regression (LR) is applied to the entire dataset to see how well a linear model has fits the CKD dataset. So, the process of adding features is stopped when addition of a new feature does not improve the LR fit on the dataset. That is how the forward selection method works. Finally, the Recursive Feature Elimination (RFE) was used to prune the least important set of features recursively from the current set of features. Similar to the forward selection, RFE also uses logistic regression to recursively eliminate features based on feature ranking.

We have implemented two different categories of feature selection process. The first category relies on the variance within features. The two methods in this category are information gain and variance threshold. The common subset of features selected from these two methods are `hemo', `pcv', `sc', `al', `sod', `pot'.

Next category of feature selection is through the use Logistic Regression model to find out which features add value to the prediction of target class "ckd" based on feature scores and ranking. Forward selection adds features if the LR model's fit improves by addition of features and RFE removes the features based on feature ranking. The common subset of features selected from these two methods are `al', `rbc', `pot', `hemo', `htn'.

Since the two categories of feature selection methods are different, hence there are differences in the selected subset of features. Combining the feature correlation, statistical significance and the different methods of feature elimination, we have 6 methods to choose the final subset of features. We would analyze if any feature falls in at least two different methods, so we would include that feature in the final list.

From the feature selection process, 13 features (`hemo', `sg', `rbc', `al', `htn', `pot', `dm', `pcv', `sc', `rbcc', `sod', `age', `bp') were narrowed down as found from the Co-relation matrix and the filter, wrapper methods of feature selection. We would discard `Sg', `Pcv', `Sod', `Pot', `rbc', `Rbcc' from the selected subset of features based on our discussions with nephrologists which is analyzed in more detail in the "Consultation with First Nephrologist" section. For our model implementation, first some of the traditional Machine Learning models were looked at. Since it's a binary classification problem, this study would start with Logistic Regression and shift towards using probabilistic models and towards more complicated model like Support Vector Machine (SVM). Next, for our explainable AI approach, ensemble tree models like XGradientBoosting (XGBoost), Decision Trees and Random Forests would be explored. The models would be fitted on the selected 6 features and then it would be observed how the models are doing in the prediction task.

\subsection{Consultation with First Nephrologist}\label{SCM}
After completing the preliminary analysis with the features, it was imperative to discuss with clinical professionals to validate the study or gain research directions. To this end, a visit was paid to Dr. Nazneen Mahmood (MBBS, MD in Nephrology) in Anwar Khan Modern Hospital, Dhaka, Bangladesh to validate whether which of these features are clinically relevant to distinguish CKD vs. Non CKD individuals. As per her findings, Specific Gravity (Sg) and Bacteria (Ba) have very little clinical significance in prediction of CKD. Moreover, the features indicating serum electrolytes like `Sod', `Pot' are not important in the diagnosis of CKD, rather they are important if any 1st line investigation provides useful analysis for further serum electrolyte analysis. Same goes for `Pcv' which is used in much later stage prediction of CKD, not in initial diagnosis. Since `Hemo' value is good enough for analyzing whether any patient has Anaemia or not, we do not separately need `rbc' and `rbcc'. In stead of these features, our access to some features like Glomerular Filtration Rate (eGFR), Bi-Carbonate (HCO3), HbA1C (Mean Blood Glucose), Pro-BNP (B-type natriuretic peptide (BNP) test), S.Ph (Sodium Phosphate), PTH (Parathyroid Hormone) etc. would be more clinically significant variables. So, efforts were established for contacting the Apollo Hospital in Madurai, India several times, specially with the contributing doctor of the dataset but to no avail. So, the scope of this research, is limited to the choice of features by choosing the ones from the feature extraction techniques discussed before and discarding the ones suggested by Dr. Mahmood. So, the final 6 features for applying the Machine Learning models are:`hemo', `sc', `al', `htn', `age', `dm'. Moreover, we discussed with Dr. Mahmood about the missing data imputation process and she also agreed that the best way of imputation would be looking at similar patients data to get estimates about the features with missing values and using just any mathematical measures like mean, median, mode would not be clinically correct.

\section{Model Application Process}
 We applied SMOTE on the imputed CKD dataset so that both the `CKD' and `Non-CKD' instances were equal. After application of SMOTE on the dataset to tackle the class imbalance issue, Stratified K-Fold cross validation is applied to divide the dataset into equal K-folds for training and testing. Stratified K-Fold is performed to eradicate the random sampling issue for the train-test split. Also, it ensures that a different sample is always chosen for the test set and no sample gets "missed out" while evaluation. For our analysis we chose k = 10, since we have 500 records after SMOTE application and with k = 10, k - 1 = 9 are the training set and 1 is the test set. So, 90\% of 500 or 450 records would be training set for each fold and 50 records would be the test set for each record. With k = 10, after all the iterations, all the records in turn would act as test set which is previously unseen to it's corresponding training set. So, the class imbalance issue is first of all tackled by using SMOTE and next by application of Stratified K-fold to get rid of random test case choosing. These methods ensure that stratified sampling is performed for model application and evaluation. The scope of this research lies fundamentally in prediciction of CKD first correctly given the independent set of variables and then application of XAI algorithms for explainability. For prediction task, the research problem is basically a binary classification task of predicting "CKD" or "Non-CKD". Hence, we used Logistic Regression first and then Bayesian Estimation or Naive Bayes model with Stratified K-Fold for train-test split.

\section{Model Performance on CKD Dataset}

Table 7 and 8 indicate the scores of the applied models in the accuracy metrics. From the previous sections on feature selection through the 4 techniques in table 6 and discussion with expert Nephrologists, final 6 features were narrowed down to compare the prediction performance of the Machine Learning models in the whole feature set versus the selected subset of features. The models discussed in the model application process are evaluated on the selected 6 features.
\begin{figure}[!h]
    \centering
    \includegraphics[scale=0.47]{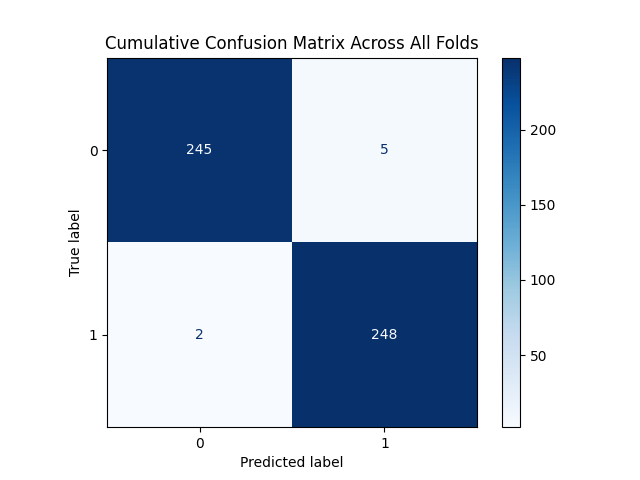}
    \caption{Cumulative Confusion Matrix for Random Forest}
\end{figure}
 We tested the model performance not just on the 4 accuracy metrics combining both classes, but also for the prediction of individual classes. We did not want to be in a scenario where the ML models performed very well in detection of one class and failed miserably in another still the accuracy scores looked really good. As evident from table 7, all the applied models have very high scores on the accuracy metrics. The number of true positives and true negatives are high and false positives and false negatives are also very low accounting for the high Precision-Recall score in all of the models. Overall, all the models have performed really well on the selected subset of 6 features for this dataset. This shows to indicate that the feature engineering process has yielded better selection of features than all combined 24 independent features. For the LR model, we chose to keep `L1' regularization parameter to penalize the wrong results by using the absolute value of the co-effecients in the LR cost function. Also, we chose "class weight" as "balanced" since after performing SMOTE on the original dataset, we have a balanced positive and negative class in the dataset. For SVM model, grid search was performed to find the best parameters suited to this dataset. For the decision tree model, "Gini Index" was used as an entropy measure to prune out branches from features since in the feature selection process we used entropy and variance to select best subset of features. Among all the models that are applied on both the entire dataset and selected subset of the features, Random Forest model has performed the best balancing all the 4 metric scores compared to all other models.

\subsection{Random Forest Model Performance}

Since of all the 7 models applied, Random Forest (RF) has performed the best in terms of accuracy metrics, this study explored more of this model performance in terms of confusion matrix, decision trees and then to the explainability.

For the Confusion matrix, we have selected to look at cumulative confusion matrix since we have used Stratified 10 fold cross validation for model evaluation. Since each fold had 50 samples in the test set, the cumulative number of samples was total 250. Looking at the confusion matrix for the Random Forest in figure 2, it is observed that 245 cases of True Positive (class 0 = CKD) and 248 cases of True Negative (class 1 = Non-CKD) prediction was found in the prediction set. On the contrary, 7 mis-classifications occurred with 2 False Positives and 5 False Negatives. Hence the Average Precision score is slightly lower than that of average Recall Score. So overall in total 250 cases of actual class of CKD, only 2 cases were there where the Random Forest model failed to correctly predict the "CKD" class and identified the two subjects as "Non-CKD" patients.

Figure 3 represents the Decision Trees generated for the Random Forests for the prediction model. The samples at the root or leaf nodes represent the percentage of samples that are reaching that node. The two values in the array represent the contribution of that feature in the prediction of Non-CKD and CKD. The root node at the top dignifies the feature value with the highest Gini Score and the next nodes at the leaves are the feature scores in terms of Gini Impurity in descending order of sample proportion. In our case, the root node starts from `al' which is indicative of the strongest feature contributor. Then the decision tree branching occurs based on gini impurity or the entropy of each feature starting with `al'. If the `al' value is less or equal to 0.587 only then the entropy value is 0.5 which is the highest for a binary classification task. At that value of `al', the decision tree would branch out to the prediction process of the two classes. If the value of `al' was higher than 0.587, the class prediction would be 0 or CKD right there. The next contributing feature is the `Hemo' and it also branches out for the two class predictions when it's value is less than or equal to 13.387. We can validate the peak value of `Hemo' towards prediction of target class by analyzing figure 10 where it is observed that at 13.387, `Hemo' is contributing most in prediction.

\begin{figure}[hbt!]
    \centering
    \hspace*{-0.8cm} 
    \includegraphics[scale=0.38]{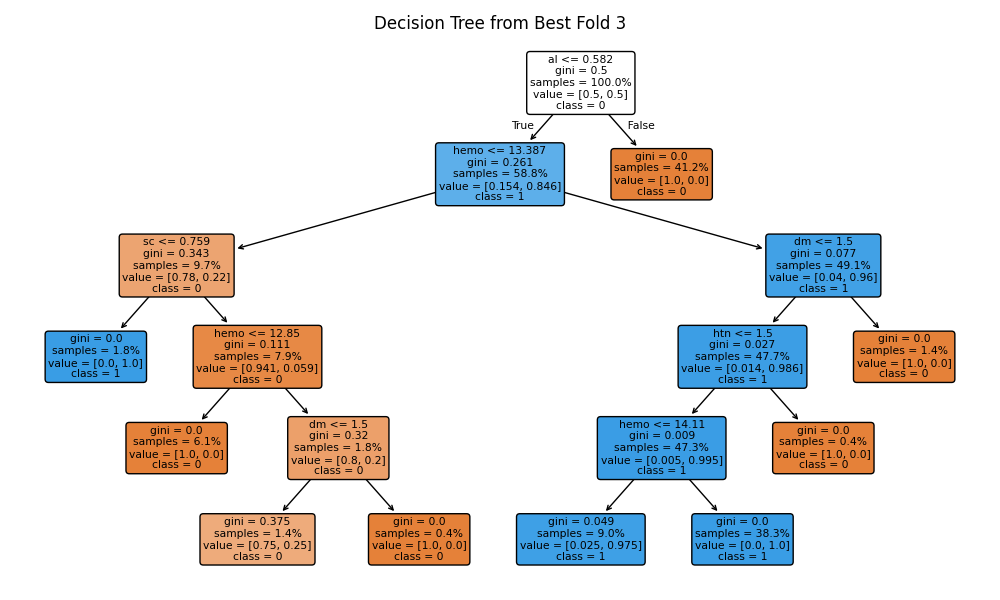}
    \caption{Decision Tree for Best Fold of Random Forest}
\end{figure}

 Point to note here is that Random Forest model combines a number of Decision Trees and based on the best pruning results of the decision trees, the final results are presented in random forest. For this study, the Grid Search technique of the Random Forest model was also employed to find out best set of parameters for prediction. It was found that a depth of 14 with number of estimators as 472, the best Random Forest Model was selected. The best performing Random Forest model would be further evaluated using the feature importances in the training phase and the validation set would be explained using XAI tools.

During the training phase as depicted in figure 4, the Random Forest model has identified the ``Hemo",``Sc", ``Al" as the features with highest weights.  The weights and coefficients of these features from the test and validation set with the help of XAI techniques were analyzed.

The LIME model was employed to explain the predictions when the prediction result is 0 and when the result is 1. The LIME is a very powerful visualization algorithm that provide us the exact range of values that the features exhibit when the prediction result is 0 and 1.

In figure 5, a random prediction case was taken where the Random Forest model predicted that the patient has CKD. The strongly contributing features for the CKD confidence are `hemo',  `al', `dm', `sc' etc.   The range of values these features exhibit were also observed when the RF model predicts as CKD. For the CKD prediction in this testcase, `Hemoglobin' values are less than 10.9. Presence of `Albumin' is indicative of CKD which is represented by `al > 2.0' here. Presence of Diabetes indicated by `Dm' value greater than 1 and high serum creatinine shown by `Sc' > 2.40.

For the Non-CKD prediction, LIME shows `al', `sc', `hemo', `dm', `htn'. Here, the LIME tool has visualized that for this instance of the patient, the best performing Random Forest model has identified it to be a Non-CKD patient because `Hemoglobin' count in gms is greater than 13.5, low level of Albumin, less chances of Hypertension, Serum Creatinine in a very low range.

\begin{figure}[!t]
    \centering
    \includegraphics[scale=0.35]{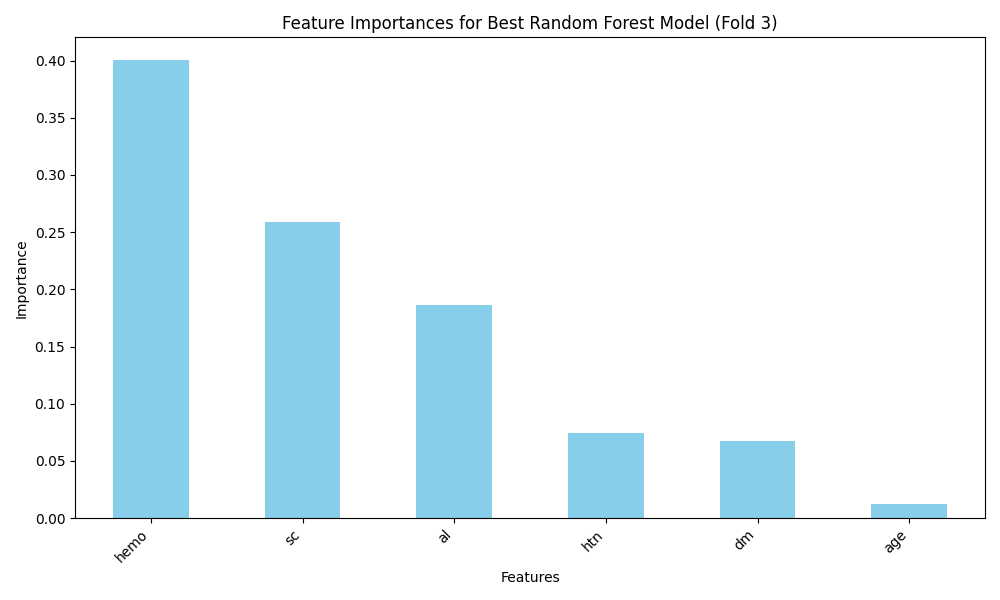}
    \caption{Random Forest Feature Importances}
\end{figure}

\begin{figure}[!t]
\large
    \centering
    \includegraphics[scale=0.42]{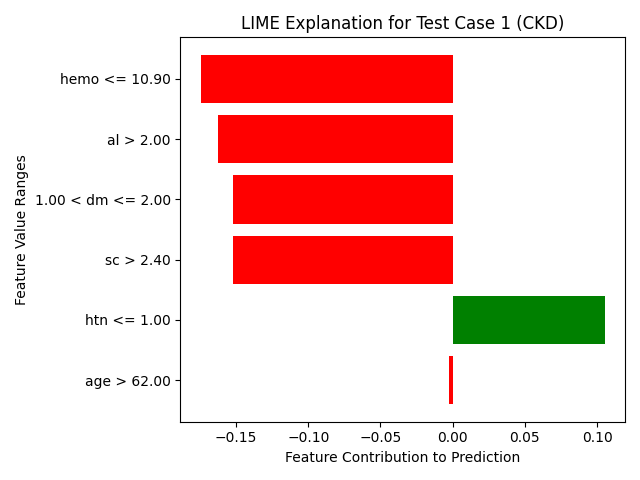}
    \caption{Random Test Case (1) Explanations by Lime}
\end{figure}

\begin{figure}[t!]
    \includegraphics[scale=0.42]{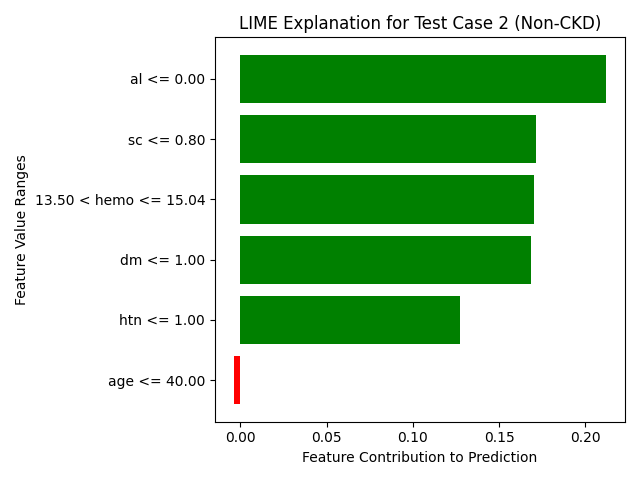}
    \caption{Random Test Case (2) Explanations by Lime}
\end{figure}

\begin{figure}[!t]
    \includegraphics[scale=0.42]{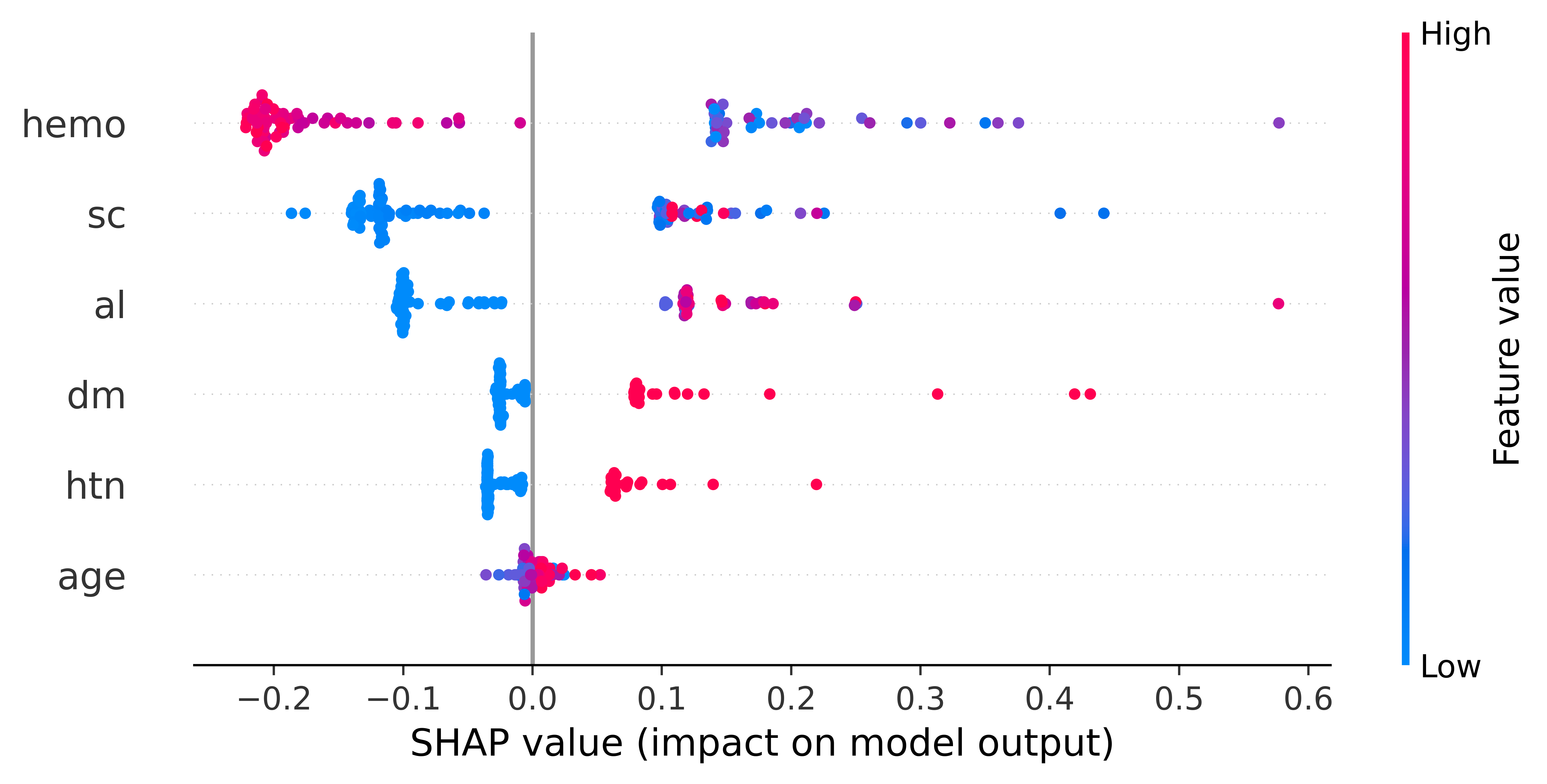}
    \caption{Shap Waterfall Plot for CKD Prediction}
\end{figure}

Model is more confident about CKD prediction when ``Sc" values are are greater than 2.40 and more confident about Non-CKD prediction when ``Sc" values are less than 0.80. Lastly, the feature ``Al" values are on the lower end for the non-CKD prediction and on the higher end for the CKD prediction as it should be as per clinical requirements.

Figure 7 shows the global impact of the features in the CKD prediction where the X-axis represents the impact score of the features and the Y-axis represents the feature values. It is observed that for high values of `hemo' and `al', generally they have high impact on CKD prediction. The impact of `age' in CKD prediction is concentrated in a region. Features like `hemo', `sc', `al' and `dm' contribute more towards the prediction of CKD compared to `htn' and `age'. Same kind of pattern is observed for Non-CKD prediction. Since SHAP provides us with the feature importances and values for the global test cases (Here we indicate the best test case chosen from 10 fold cross validation) unlike LIME, the two figures represent the feature importance scores and the feature values exhibited by the independent features for the contrasting class predictions.

Fig 8 depicts the scatter plots of global test cases to visualize the inter-dependence of independent features. Here, the top two features with respect to feature contribution i.e. ``al" and ``hemo" are analyzed. Presence of two clusters could be noticed in the inter-dependence plot where `hemo' values range from 8 to 13 and `al' values range from 1.7 and above. The 2nd cluster is for `hemo' values ranging from 14 to 18 and `al' values from 0 to 1.0. The two clusters indicate CKD vs Non-CKD prediction where the first cluster represents the CKD predictions for the test cases and the 2nd cluster represents the Non-CKD predictions.
 
\begin{figure}[!t]
    \centering
    \includegraphics[scale=0.5]{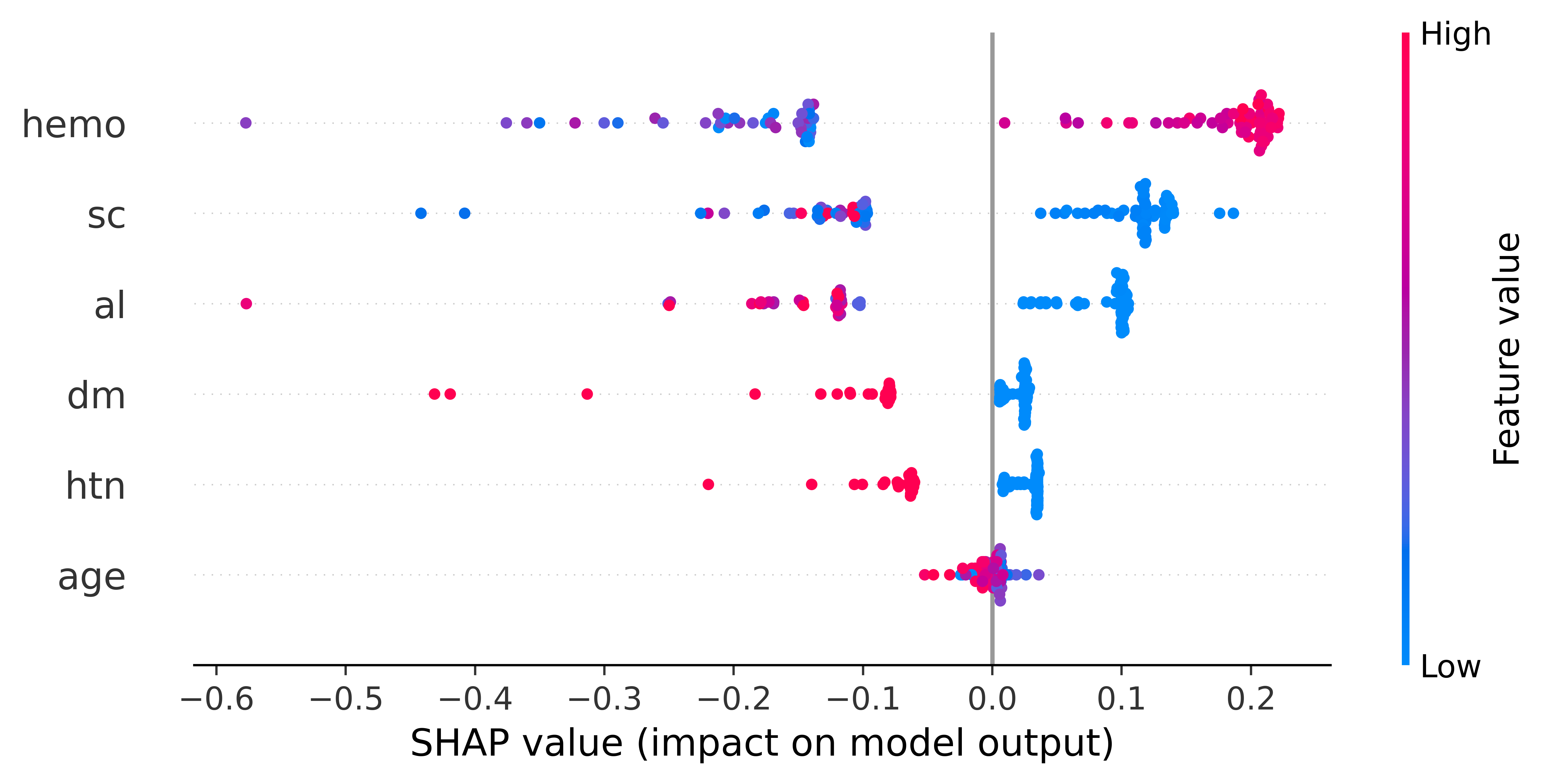}
    \fontsize{18}{12}\selectfont
    \caption{Shap Waterfall Plot for Non-CKD Prediction}
\end{figure}

The peak values of prediction are also validated by the Accumulated Local Effects (ALE) plots in fig 10. In the X-axis, Haemoglobin level are plotted against the feature contribution scores in the Y-axis. It is seen that the feature contribution for `Hemo' is the highest when the `hemo' value is close to 13 and the contribution score remains constant with the increasing values.

\begin{figure}[!t]
    \includegraphics[scale=0.5]{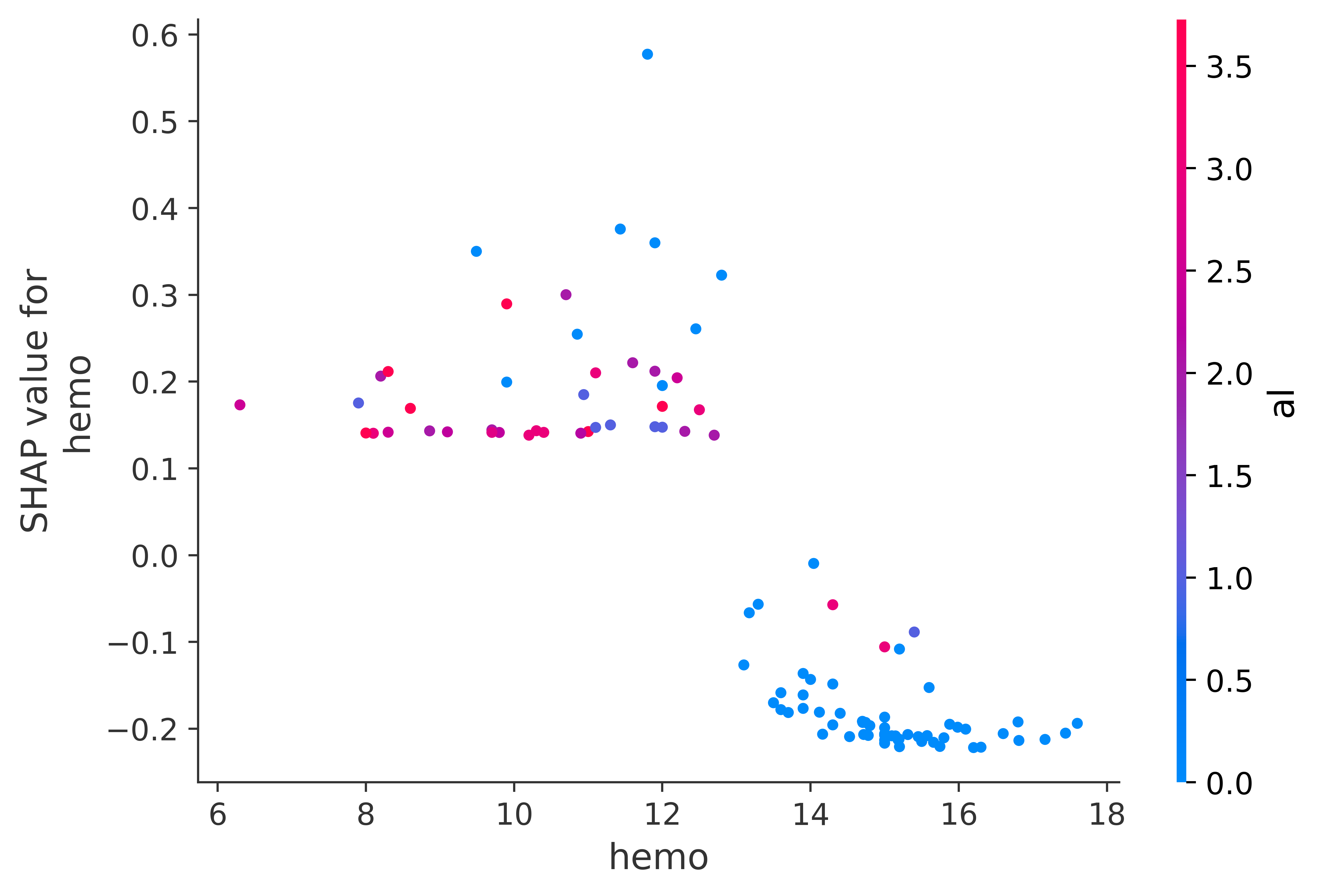}
    \caption{SHAP Inter Feature Dependence Plots for Global Test Cases}
\end{figure}

\begin{figure}[h!]
    \includegraphics[scale=0.45]{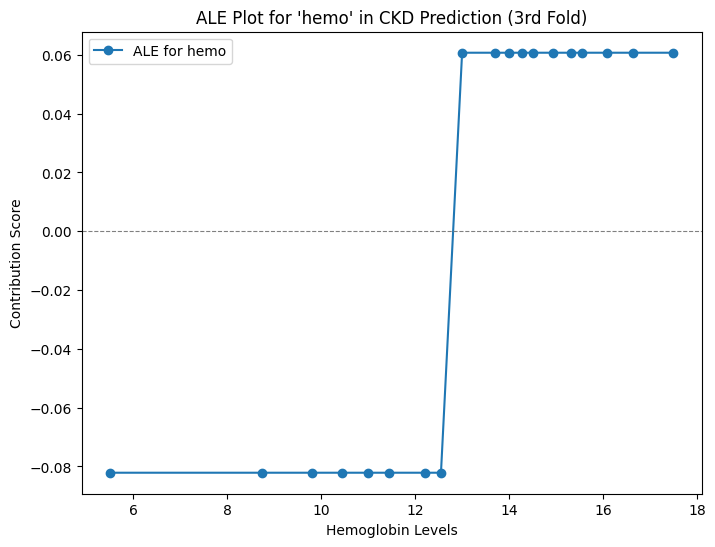}
    \caption{ALE Plot showing effect on prediction using 'Hemo'}
\end{figure}

\begin{figure}[h!]
    \includegraphics[scale=0.24]{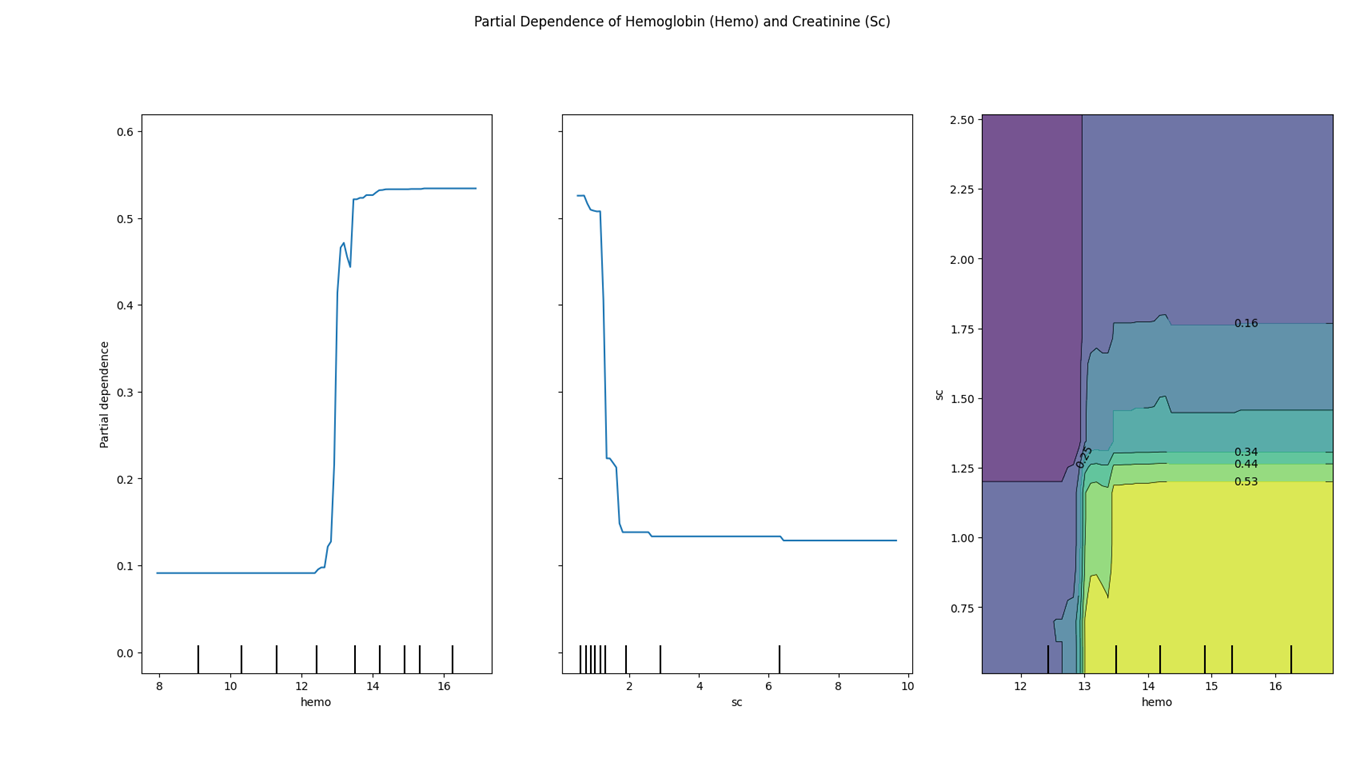}
    \caption{PDP Plot showing effect on prediction using 'Hemo' and 'Sc'}
\end{figure}

The Partial Dependence Plots (PDP) in figure 11 validate how the feature values dependency works in the predictions. If the partial dependence plot in fig 11 is analyzed, it would be seen that the more strongly contributing features `hemo' has their peak contribution when `sc' values are greater than 1.25 and `hemo' values are greater than 15. `Hemo' values also reach a local maxima at values 12.5 to 13. The combined dependence of the two features is also shown when local maxima is reached at 0.44 and 0.34 (in terms of contribution towards model prediction) when the `Hemo' value is greater than 15, `Sc' is greater than 1.25 and another at 0.55 when the `Sc' is less than 1.25 and `Hemo' value greater than 15.

\subsection{Example Based Techniques and Implementations}
For the example based techniques, Counterfactuals and Contrastive Explanation Models (CEM)s are to be applied to this study. From the Counterfactuals, a specific set of rules would be found out based on the prediction. For the Contrastive Explanation Models, local prediction in a test case would be looked at and then the CEM tool would tell us what changes would need to happen for a subset of the selected features for the best performing Random Forest model to predict the contrasting class.

\begin{figure}[hbt!]
    \centering
    \includegraphics[scale=0.35]{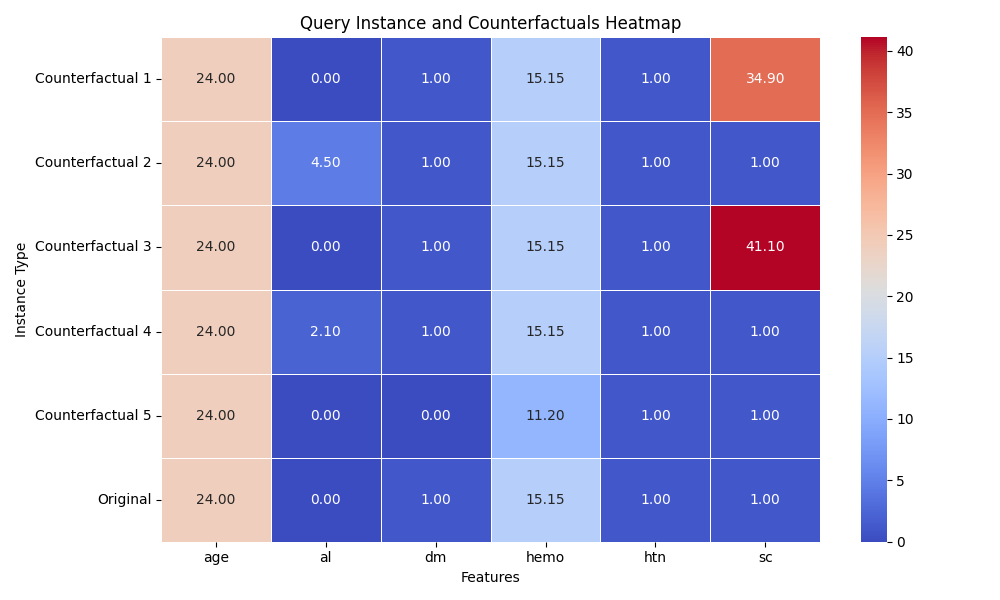}
    \caption{Counter Factual for the Original Prediction of `Non-CKD' versus the contrasting `CKD' predictions}
\end{figure}

From fig 12, the last row ``Original" represents the independent feature values for a Non-CKD prediction for a 24 year old patient. The other 5 counterfactuals in the rows above indicate the corresponding changes for each feature value for the contrasting class `CKD' prediction. It is seen from counterfactual 1 and 3 that high value in `Sc' or Serum Creatinine is present for CKD prediction. Also, the `Albumin (Al)' quantity is high for the 2nd case of CKD prediction. But if we focus on the last row `Original' prediction of Non-CKD we can observe that, in order for the model to predict the Non-CKD case, the independent feature value for `Al' is very low and `Sc' which is the fundamental bio-indicator as clinical marker for CKD are also very low. So if this heatmap is presented to a physician, the case of CKD examples could be presented and informed that, to prevent the CKD from progressing to the next stages these are the factors that need to be further investigated upon.

\subsection{Interpretability Measures of Ensemble Models}
Since primary objective of this research was to apply the Explainable AI tools for prediction and analysis of CKD variables, it is also important to analyze how the Ensemble Models applied here perform on certain measures like Interpretability, Fidelity, Fidelity Interpretability Index (FII) and Fidelity Accuracy Index (FAcc). FII signifies the combined score of the model in both fidelity and interpretability. The higher the score, the more trust we can put in the model for it's decision results and predictions. Also, we can explain the decisions in an easier way if the FII score is higher for a model. FAcc signifies a balanced score of Fidelity with the Actual Accuracy of the model. Since we want both accurate model that we could trust, that is why first of all we need the model predictions to be accurate and then look for trust in the models decision process. First we would calculate the Interpretability (I). Studies conducted by Thomas Norrenbrock et al at \cite{b67} have established that the fewer the number of features chosen while evaluation, the easier it is to interpret the model decision-making to the laymen. To find the interpretability score, we would first find out the feature importances in the test set of the best fold in the 10 fold cross validation of the Random Forest, Adaboost and Xgboost model. We would also employ SHAP as a surrogate model to find out feature importance scores across the global
test cases. Then use a similarity measure i.e. Cosine Similarity to find out how similar the feature contribution scores are for the two different models. The cosine similarity in equation (3) is specified for our use case by changing the original `u' and `v' variables. The reason for us to use this similarity metric is, we don't just want to rely on ensemble models feature importance, rather we want to validate with SHAP for correctly estimating number of important vs. redundant features. To find a threshold for feature importance, we would rather employ the cumulative feature importance scores and discard features that contribute to less than 10\% on the cumulative feature importance scores. For the cosine similarity in equation (3), EM represents the feature importance score from the model and LIME represents the feature importance score by LIME explanations.

\begin{equation*} I= \frac{\mathrm{Number of Redundant Features}}{Actual Number of Features} \tag{1}\end{equation*}

Looking at the feature importance scores of the Xgboost in figure 13 and figure 15 by LIME, it is observed that 4 features are contributing more than 90\% of the cumulative feature importance scores. Also, they (`al', `dm', `hemo', `sc') have a high cosine similarity score 0.82 with the LIME feature importance explanation scores. So, these 4 features are deemed as important by the XGBoost and LIME global feature explanation importances. Hence, the number of redundant features are (24-4) = 20 for the XGBoost model. So, as per equation (1), the Interpretability score for Xgboost model would be 20/24 = 0.83.

For the Adaboost model, based on figure 13 and 16, the Cosine Similarity scores are high only when features `sc' and `age' are chosen. But `hemo' contributes significantly for the cumulative feature scores in it's own feature rankings. Hence, the number of important features for interpretability are 3. So, the number of redundant features would be (24-3) = 21. So equation (1) follows the Interpretability score for Adaboost = 21/24 = 0.87.

For the Random Forest model, the LIME explanation scores are similar to that of Adaboost. So, looking at figures 4 there are 4 features `hemo', `sc', `al', `htn' that contribute towards more than 90\% of the cumulative scores. But in figure 16 of the LIME explanations, we see only `hemo' and `sc' would result in a high cosine similarity score and the aggregate cosine similarity would reduce if we include `al' and `htn' in the equation 3 for calculating cosine similarity. Hence, Interpretability score for Random Forest is 22/24 = 0.91. 

To calculate Fidelity score in the Xgboost we are mainly calculating the external fidelity score. External Fidelity score is used to develop an explanation approach for the underlying predictive ensemble models given it's equivalent interpretable model. The ensemble models Random Forest, Xgboost and Adaboost all rely on different Decision Trees to make their predictions. Hence, it is easier to evaluate the explainability of these ensemble models with the help of their corresponding Decision Trees. The Decision Trees could be thought as white box models capable of explaining the predictions of the black box ensemble models to the end users. The scope of using external fidelity measure is used in many existing concurrent research, specially in \cite{b68, b20}. The external fidelity would be calculated using the Precision and Recall but in context of fidelity. In this context, true features are the ones selected by the ensemble model and explanation features are the ones selected by white box Decision Tree model.

\begin{equation*}
R = \frac{\mathrm{True\ Features} \cap \mathrm{Explanation\ Features}}{\mathrm{Explanation\ Features}} \tag{2}
\end{equation*}

\begin{equation*}
P = \frac{\mathrm{True\ Features} \cap \mathrm{Explanation\ Features}}{\mathrm{True\ Features}} \tag{3}
\end{equation*}

\begin{table}[!t]
\centering
\caption{Feature Score in Test Set as Found by the Decision Tree}
\begin{tabular}{|c|c|}
    \hline
    \textbf{Feature Name} & \textbf{Importance Score} \\ \hline
    hemo & 0.788874 \\ \hline
    al   & 0.114317 \\ \hline
    dm   & 0.049884 \\ \hline
    sc   & 0.034721 \\ \hline
    age  & 0.012204 \\ \hline
\end{tabular}
\end{table}

The precision score and the recall score would be calculated using the equations (3) and (4). The F1 score as the harmonic mean of the two scores would be calculated as the final External Fidelity Score for the ensemble model.

It is seen from table 8, that only top two features are accounting for 90\% of the cumulative feature scores, hence these two are our explanation features. For the true features in Xgboost in figure 14, it is evident that 4 features are contributing for the cumulative scores, hence the number of true features are 4. So, the Precision and Recall of Xgboost in terms of Fidelity are 2/4 and 2/2 respectively. Similarly we can calculate the scores for the other two ensemble models and finally calculate F1 as the measure of external fidelity. The final scores of interpretability and fidelity are populated in table 9 for all the models with their FII and FAcc values.

For the Adaboost model, true features are 3 and for the Random Forest model, it is 4. Hence, the Precision score for Adaboost is 2/3 and 2/2 respectively. For Random Forest, it would be 2/4 and 2/2 respectively. 

\begin{figure}[!t]
    \centering
    \includegraphics[scale=0.4]{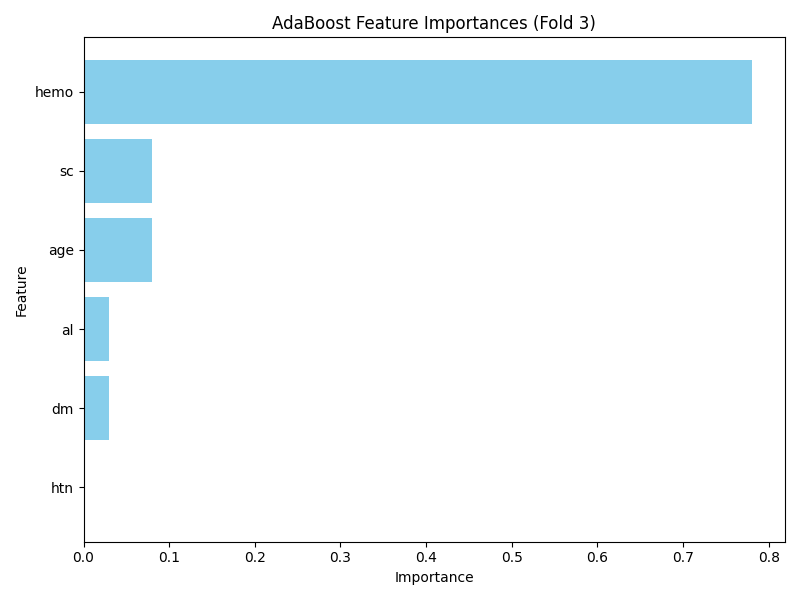}
    \caption{Feature Importances in Test Set for Adaboost}
\end{figure}

\begin{figure}[!t]
    \centering
    \includegraphics[scale=0.45]{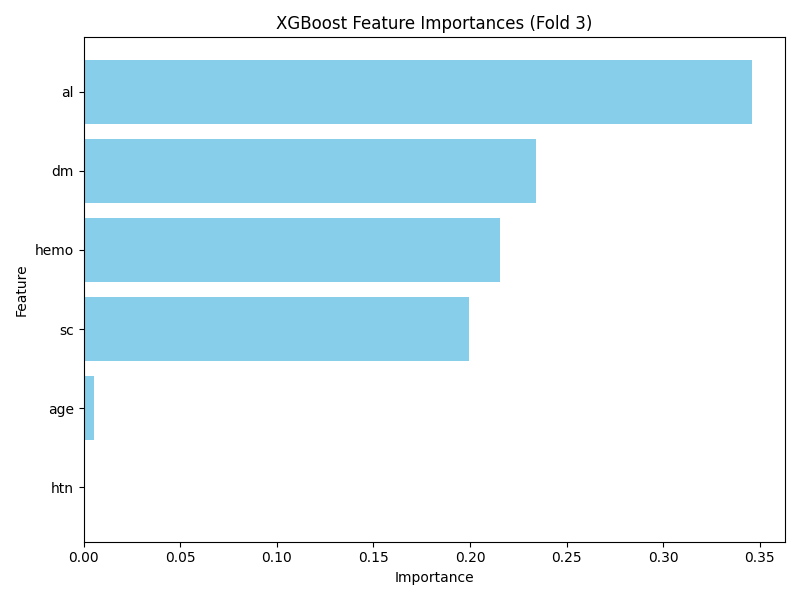}
    \caption{Feature Importances in Test Set for Xgboost}
\end{figure}

\begin{equation*}
\text{Cosine Similarity} = \frac{\sum_{i=1}^n (\text{EM}_i \cdot \text{LIME}_i)}{\sqrt{\sum_{i=1}^n (\text{EM}_i^2)} \cdot \sqrt{\sum_{i=1}^n (\text{LIME}_i^2)}} \tag{4}
\end{equation*}

The other two metrics in the interpretability are the Fidelity Interpretability Index (FII) and Fidelity Accuracy Index (FAcc). These metric scores are calculated for all the 3 ensemble models: Random Forest, Adaboost and Xgboost.

\begin{equation*} FII= \mathrm{F}\times I\tag{5}\end{equation*}

\begin{equation*} FAcc= \mathrm{F}\times Acc\tag{6}\end{equation*}

\begin{table}[!h]
\centering
\caption{Interpretability Measures In the Test Set of Ensemble Models}
\begin{tabular}{|l|c|c|c|c|}
\hline
\textbf{Model Name} & \textbf{Interpretability} & \textbf{Fidelity} & \textbf{FII} & \textbf{FAcc} \\
\hline
AdaBoost & 0.87 & 0.80 & 0.70 & 0.77 \\
\hline
Random Forest & 0.91 & 0.67 & 0.61 & 0.66 \\
\hline
XgBoost & 0.83 & 0.67 & 0.55 & 0.66 \\
\hline
\end{tabular}

\end{table}

\begin{figure}[!t]
\centering
    \includegraphics[scale=0.4]{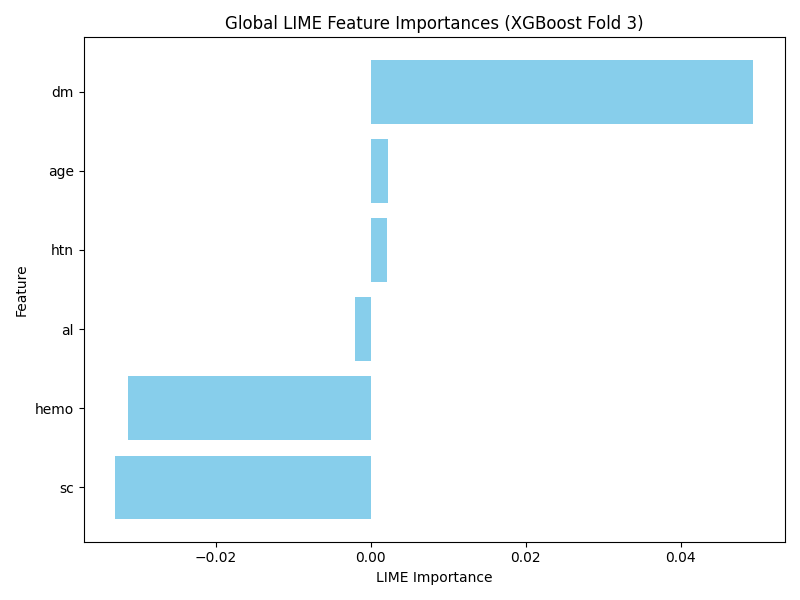}
    \caption{Feature Importances in Test Set for Xgboost Using LIME}
\end{figure}

\begin{figure}[!t]
    \centering
    \includegraphics[scale=0.4]{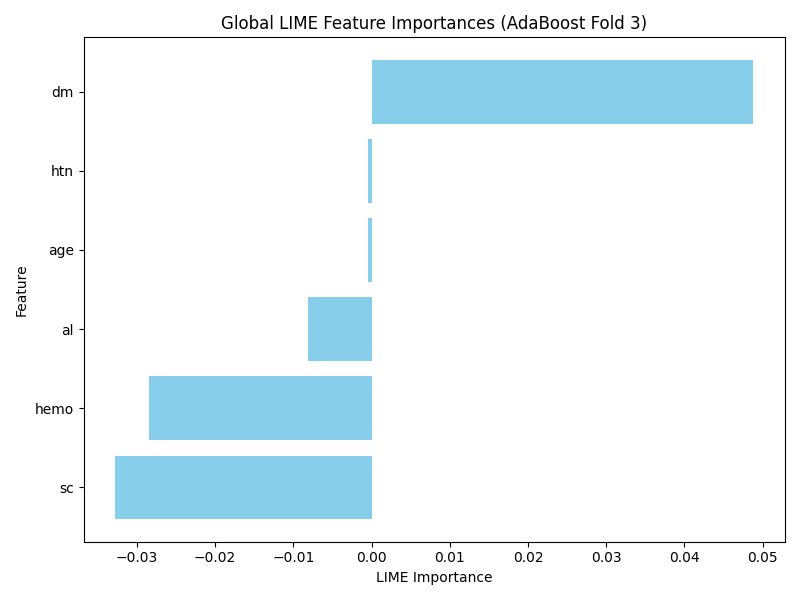}
    \caption{Feature Importances in Test Set for Adaboost Using LIME}
\end{figure}

\subsection{Comparison with Existing Research}
Exisiting research on the applications of Explainable AI techniques have mostly focused on Feature Based Techniques like Shapley Additive Explanations (SHAP). To this end, study conducted by Gabriel et al is a quite comprehensive study where they employed Case Based Reasoning (CBR) approach to identify causation of CKD from case by case basis on Colombian population \cite{b21}. Similar to the work conducted \cite{b22} on investigation of XAI models on CKD from patient dataset gathered from a local hospital in Taiwan, Local Interpretable Model Agnostic  Explanations (LIME) were employed to explain the predictions of the models for each cases of CKD or non-CKD. Most of the XAI feature based tools were used like Partial Dependence Plot (PDP) and Accumulated Local Effects (ALE) to validate which features contribute in which particular value range and also to analyze the inter-feature dependence in contribution scores.

The novel feature of this study is the use of Example Based Techniques like CounterFactuals and Contrastive Explanation Models (CEM) to explain what changes needed to happen to each feature for a change in the outcome variable. To the best of our knowledge, no published work so far has performed similar analysis on the tabular dataset of this CKD UCI ML one.

The Fidelity Interpretability measures were also employed to compare our study with that of \cite{b20} and we found out that both the Random Forest and Adaboost model has performed considerably better than theirs in the interpretability score. Point to note here is that we have used a more thorough approach of estimating interpretability score using cosine similarity and cumulative feature scores. Also, the process of estimating Fidelity score using Precision Recall and F1 in our ensemble models is quite different than them using just the accuracy of a interpretable decision tree classifier with the accuracy of the ensemble model. The different measures to calculate interpretability and fidelity have resulted in difference of values in the FII values where our Adaboost model significantly outperforms theirs but Xgboost model has almost 23\% lower score.

The difference of this research with that of \cite{b20} is that this research has eradicated features like ``Specific Gravity (Sg)" from our analysis beforehand after consulting with a nephrologist in Bangladesh where her feedback was, they don't consider at any stage the Specific Gravity to be an important factor for CKD prediction. Moreover, features like Pus Cell Count (Pcc) has also been removed as this feature is mostly used for determining Acute Kidney Injury (AKI) which is not the current scope of our research. Also, we have removed features like `Puss Cell Volume (Pcv)' because of it's effectivity in identifying later stages of CKD. Features like `Rbc', `Rbcc', `Wbcc' are removed because `hemo' already provides the clinical insight on the blood cell levels required for CKD prediction. On the data pre-processing stage, the difference of our study with the concurrent ones, lie in analyzing the pattern of missingness of data, imputation through learned representations, validating the statistical significance of features with feature co-relations, employing several techniques for feature elimination, employing example based XAI algorithms, calculating interpretability with cosine similarity and calculating external fidelity to compare with concurrent research. So, to sum up, validating all the decisions through multiple measures in the pre-processing to model application to model performance analysis, direct consultation with multiple nephrologists and modifying the study with their feedback etc. are the new scope of work performed in this study different from the existing ones.

\section{Discussion on the Results}
The primary focus of this study is to analyze the Machine Learning models decision making process in terms of feature contribution, range of feature values, peak and constant contributions, reasons of choosing a certain prediction, changes needed from feature ends for a contrasting class prediction etc. To this end, the Interpretability Fidelity measures were employed. Analyzing in terms of accuracy metrics and interpretability measures, the Adaboost model performs better than all other models specially when Fidelity Interpretability Index (FII) is considered. But the Random Forest model's performance can't also be ruled out as it has higher accuracy, F1 score and Interpretability value. So, from a clinical perspective both the models are presentable to the clinicians where the Random Forest model would provide a few less features as important contributors for prediction than Adaboost model.

This study focused more on the interpretability of the models with explainable AI to make the decision-making process of the model easier to understand for the general people. All the techniques like LIME, SHAP, ALE have validated the feature contributions, inter-feature dependence, feature values in CKD vs Non-CKD predictions. For this scope of work,  both the feature-based and example-based explainable AI techniques were focused. To our knowledge, no research has yet implemented the example-based techniques like Counterfactuals or Contrastive Explanation Models (CEM) for the domain of using Explainable AI in Chronic Kidney Diseases. Also, the interpretability and fidelity is not measured in any concurrent research on CKD with XAI as ours. If the ML models do well in terms of interpretability, it makes the job of the researchers, clinicians and the general people easier on all ends to understand the prediction models.

\section{Limitations of Current Dataset and Potential Improvements}

 From the dataset perspective, this study is dealing with a CKD dataset containing 250 patients with CKD and 150 patients not having CKD. This dataset collected from Apollo Hospital, India is just focused on one ethnic group for this study. So naturally this dataset lacks diversity in demographic variables like ethnicity, gender and socio-economic aspects. Since the patient distribution was organized like having CKD patients occurring consecutively and similarly non-CKD patients, dataset was shuffled to make them distributed. Presence of a significant portion of missing values in a sensitive domain like CKD is also a great limitation of this dataset. As we had to do thorough research starting from the statistical persepective of first analyzing the pattern of missingness, then employ feature engineering and consultation with our first nephrologist to use the learnt representation from observed data to impute the missing values. Also, from the Pearson Correlation, it is found very strong correlation coefficients among features which was also validated through statistical measures on the Logit model. Presence of strong inter-feature correlation only makes the job of imputation harder. Moreover, there was strong positive variation in the correlation accounting for high accuracy scores. These factors have played important roles in the prediction results of the ML models. So, the limitations of the dataset and their impact on the findings are discussed in a concise list:
 \begin{itemize}
     \item Large amount of missing values and high inter-feature correlation make it difficult to pre-process the dataset specially, creates a dilemma of keeping and discarding independent features.
     \item Since the dataset lacks information in demographic variables, a thorough analysis of the demographic variations could not be performed which is important for clinical research.
     \item Lacking of information in the more useful clinical investigations makes the clinical practitioners ask critical questions about specific patient findings.
 \end{itemize}

\subsection{Discussion with Second Nephrologist}
 We also had our one to one session  Dr. Md. Nabiul Hassan Rana in Asgar Ali Medical College, Dhaka, Bangladesh. Both of the nephrologists (Dr. Nazneen Mahmood previously) informed us that the current dataset needs more variables for a conclusive prediction of the CKD patients. Based on the class distribution in the dataset, some of the records does not really contain conclusive reasons for declaring a patient having CKD. For example, a 24 year old patient with only history of Diabetes Mellitus had been declared having CKD without any other discriminative feature values. It is unusal for a patient of 24 years to develop CKD so that's where their suspicion arose. Moreover, another patient with high Albumin presence was declared having CKD without any other conclusive feature values. The high protein in the urea of that patient could happen for a multitude of reasons like heavy exercise, steroid, excess protein intake etc. Some of the features like `Pcc' is used much later to track the progression of CKD to AKI, not to predict the cases of CKD. Features like ``Sg", ``Bgr" are not important to predict the cases of CKD. The large percentage of missing data in a lot of features is also a critical aspect to this dataset. To address the dataset issues we tried to get in touch with the contributor of the dataset in Apollo Hospital, Medurai India, but to no avail. So, the research contributions made by these nephrologists are:
 \begin{itemize}
     \item Guiding us towards the correct means of missing data imputation for the imputed data to have clinical relevance.
     \item Analyzing the existing set of features and helping us in the feature elimination methods (removing `Sg', `Sod', `Pot', `Pcv', `Rbc', `Rbcc' from the selected features) so that selected subset of features on this dataset provide clinical significance in CKD prediction.
     \item Aiding us in analyzing the potential limitations in the feature values and the diagnosis results of this CKD dataset in terms of ethnicity, proper investigations and results, associated disease, prognosis and complications etc.
     \item Assisting us in collection of data from patient visits to Asgar Ali Hospital in terms of the aforementioned category of variables.
 \end{itemize}
 
\subsection{Approach to developing a new dataset}
To extend the scope of our research, this study collaborated with Nephrologists and Intern doctors at Asgar Ali Hospital in Dhaka, Bangladesh and they have agreed to share their CKD patient data with us. From their feedback on the current research and potential improvements, a list of features which are distinguishing factors for CKD prediction and would contain more information than the current CKD dataset was prepared.

\begin{table}[!t]
\centering
\caption{New Dataset Potential Features}
\begin{tabular}{|p{4 cm}|p{4 cm}|}
	\hline
	\textbf{Domain} & \textbf{Details} \\
	\hline
	Demographic Variables & Age, Ethnicity, Family History, Drug History, Smoking History, Obesity \\
	\hline
    Associated Disease & Diabetes Mellitus, Hypertension, Cardiovascular Disease, Systematic Inflammatory Diseases, Congenital Disease\\
	\hline
	Clinical Data & Nocturia, Oliguria, Blood Pressure, Appetite, Itching, Pedal Edema, Breathing Difficulties\\
	\hline
    Biochemical Variables & Serum Creatinine, Blood Urea, eGFR, Urine R/E, Urine for ACR, Serum Electrolyte\\
\hline
Prognosis \& Complications & Iron Profile, HbA1C \& FBS, Lipid Profile, Bone Profile, Renal USG \& Imaging\\
\hline
\end{tabular}

\end{table}

As seen from table XI, the feature set is prepared after long consultation with nephrologists working with CKD patients for more than 15 years, so the positive findings would be found from Associated Disease, Clinical Data, Bio-Chemical Variables, Prognosis and Complications and generate a more informative dataset on which the Machine Learning models would be applied later.

\subsection{Working with Security Aspects of Dataset}
Since the current CKD dataset is already generated from different blood and urea tests of patients, there is little to no scope of working with generated data. But in the process of collection of new data, this study is looking towards prevention of any Poisson attacks on the dataset by ensuring the confidentiality in the transfer of information from the hospital to us. The application and saved model performances would be limited to our systems only in the S.M.A.R.T lab of Wright State University so there would be no scope of Membership Inference Attacks. The confidentiality ensured from both the ends of the hospital to the researchers at the university would prevent any Sybil or Data Extraction Attacks on the newly collected dataset.

\section{Conclusion}
This study has focused on XAI tools towards developing Machine Learning models that not only perform well in the CKD prediction problem, but also are interpretable to the physicians as end users. Our research goal from this study was to identify specific set of variables that contribute closely towards identifying the CKD patients and make strong enough cases for distinguishing CKD patients and Non-CKD healthy individuals.  The value ranges of these variables were shown with LIME, showed the peak contribution using ALE and PDP plots, global contribution scores using SHAP, necessary changes in the features through CounterFactuals, conditional clauses for CKD in stead of Non-CKD prediction and vice versa using CEM. The models were compared in terms of interpretability measures with that of concurrent research works in CKD with XAI. With the limited dataset, this study resorted towards developing a more informative dataset from a well reputed hospital in Bangladesh where the scope of this study would be enhanced to practice the security means to prevent the new dataset against any adversarial attacks. The research problem is formulated with XAI tools so that the physicians as end users can gain insight into the causes of CKD for patient by patient basis and provide appropriate medications or life style changes so that the CKD from stage 1 does not progress to later stages. The later stages of CKD can bring huge costs financially and healthwise specially if the patient develops End Stage Renal Disease (ESRD) where dialysis remains their only option. In a lot of 3rd world countries these CKD prediction is not detected till the 3rd or 4th stage which is much more dangerous for them. Moreover, in a lot of public hospitals there is no scope of treatment in the final stages and many of them would not be able to afford private hospitals for treatment. That is why this research problem is highly crucial for us to analyze and present our findings from all angles of features, information and security to help the clinicians and the general people as much as possible.

\EOD
\end{document}